\documentclass{article} %
\usepackage{iclr2027_conference,times}

\usepackage{amsmath,amsfonts,bm}

\def\eqref#1{equation~\ref{#1}}

\def\1{\bm{1}}

\DeclareMathAlphabet{\mathsfit}{\encodingdefault}{\sfdefault}{m}{sl}
\SetMathAlphabet{\mathsfit}{bold}{\encodingdefault}{\sfdefault}{bx}{n}

\usepackage{hyperref}
\usepackage{url}
\usepackage{amsmath,amssymb}
\usepackage{graphicx,booktabs,multirow,array}
\usepackage[table]{xcolor}
\usepackage{placeins}
\usepackage{wrapfig}
\usepackage{bbding}
\definecolor{lightgreen}{RGB}{53,193,141}
\definecolor{darkred}{RGB}{192,0,0}

\definecolor{baseCell}{HTML}{FAE9E9}   %
\definecolor{oursCell}{HTML}{EDF5EA}   %

\title{Tell Robot What Not to Do: A Negation Understanding Perspective}

\author{\parbox{\dimexpr\textwidth-2\tabcolsep\relax}{\centering
\normalfont Fazeng Li\textsuperscript{1,*}, Gan Sun\textsuperscript{1,\ensuremath{\dagger}}, Hao Cheng\textsuperscript{1,*},
Weihong Ren\textsuperscript{2}, Yang Cong\textsuperscript{1}\\[5pt]
\small \textsuperscript{1}South China University of Technology\\
\textsuperscript{2}Harbin Institute of Technology\\[3pt]
\footnotesize \textsuperscript{*}Equal contribution.\quad
\textsuperscript{\ensuremath{\dagger}}Corresponding author.
}}

\iclrfinalcopy %
\begin{document}

\maketitle

\begin{abstract}
Instruction following enables robots to perform diverse tasks specified in natural language, making it a fundamental capability for human--robot interaction. 
Beyond communicating desired outcomes, users also need to specify constraints on \emph{what not to do}. 
We investigate how to enable vision-language-action models (VLAs) to follow negated instructions, where robots must accomplish task goals while respecting explicit exclusions.
To this end, we propose NegaAlign, a parameter-efficient, plug-and-play framework that extends pretrained VLAs to follow negated instructions through image--language supervision alone. 
Specifically, we introduce Negation Transformation Layers into selected layers of the vision--language backbone to reshape intermediate instruction representations. 
Meanwhile, teacher-guided alignment mechanism is designed to align the instruction-relevant visual tokens, transferring action-relevant grounding from instructions that satisfy the negated constraint. 
Training phase uses supervision constructed from existing demonstrations and updates only the inserted layers, keeping all pretrained parameters frozen, including the action generator.
We further introduce NegaBench, a simulation benchmark spanning 10 scenarios across five domains for systematically evaluating manipulation under negated constraints. 
Experiments across GR00T, $\pi_0$, and $\pi_{0.5}$ demonstrate consistent improvements in negated instruction following. 
With 11.6M trainable parameters, NegaAlign increases the negated-instruction success rate of $\pi_{0.5}$ from 2.60\% to 88.45\% on NegaBench and from 12.4\% to 88.8\% on real-world tasks, while retaining performance on affirmative instructions.
\end{abstract}

\section{Introduction}
\begin{wrapfigure}{r}{0.55\textwidth}
    \centering
    \includegraphics[width=0.5\textwidth]{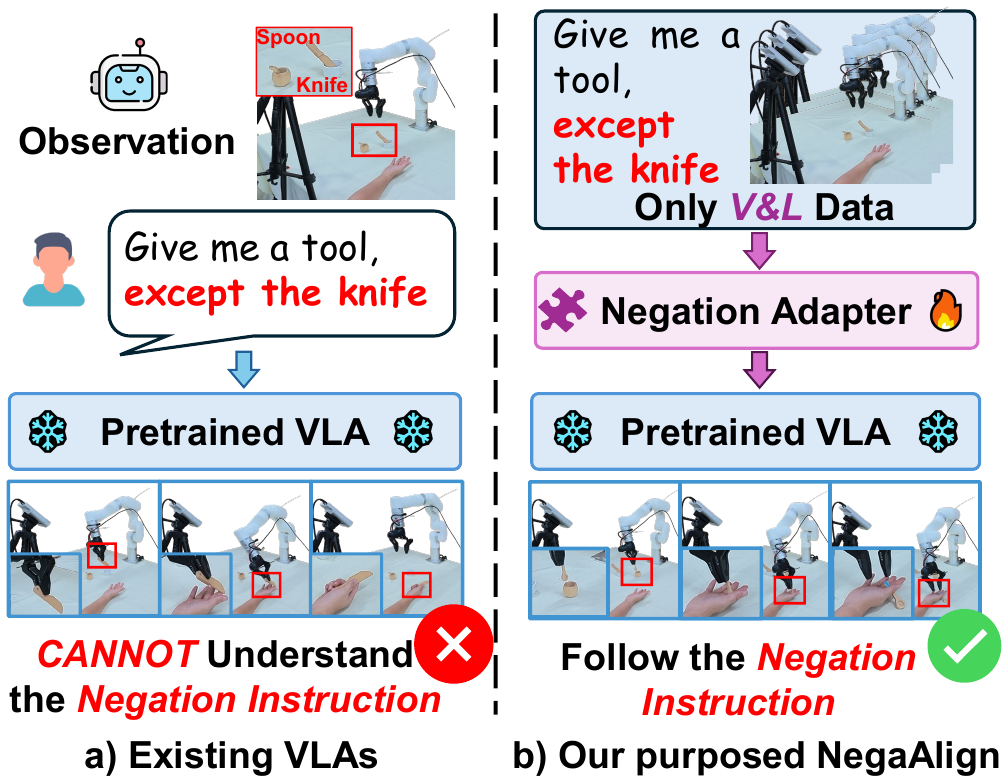}
    \caption{Existing VLAs struggle with negated instructions, risking unsafe actions. Our proposed NegaAlign enables actions under negated constraints.}
    \label{fig:teaser}
\end{wrapfigure}
As robots enter daily life, reliable instruction following is crucial for both human and robot safety. Vision Language Action models (VLAs)~\cite{zitkovich2023rt, pmlr-v270-kim25c, intelligence2025pi05visionlanguageactionmodelopenworld, bjorck2025gr00t} follow natural language instructions and perform manipulation tasks in complex environments. Pretraining on large-scale embodied datasets~\cite{10611477} enables them to complete a wide range of manipulation tasks, including pick-and-place, tool use, and object rearrangement.

However, these existing VLAs are mostly trained on affirmative instructions and struggle to follow negated instructions. 
This limitation can result in unsafe or undesired actions~\cite{chen2026hazardarena, lyu2026foresightsafety, zhang2025safevla}, as the robot may act on the affirmative content of the instruction without considering the negated constraints. 
\textit{For example, given \textit{"Give me a tool, except the knife"}, a robot will still hand the knife to the user, which is an extremely dangerous behavior in daily life, as illustrated in Figure~\ref{fig:teaser}(a).}
A key source of this problem lies in the VLMs backbone, whose difficulty in understanding negation affects the action sequence generation of VLAs further.
Existing studies investigate negation understanding in image-text matching \cite{dobreva2021investigating, aggarwal2026seeing}, image editing \cite{Bui_2025_CVPR}, and image-grounded object detection \cite{kang2026what, yang2026mastering}, which could not be applied to the VLAs directly since the VLAs need to ground the negated instruction into the sequence of actions to complete the task.
Recent approaches improve VLA instruction following through hierarchical decomposition and fine-grained supervision~\cite{zawalski2024robotic, 10611525}.
Hi Robot \cite{shi2025hirobotopenendedinstruction} uses an additional VLM to decompose instructions into subtasks, introducing extra inference computation. 
APT~\cite{xu2026aptactionexpertpretraining} adopts two-stage pretraining to better preserve VLM knowledge in the action expert, but requires a dedicated multi-stage training procedure, while Fine-VLA~\cite{hu2026finevlafinegrainedinstructionalignment} constructs fine-grained instruction data to strengthen language-action alignment.
However, collecting negation-specific robot demonstrations is costly and time-consuming.
Hence, how to enable existing VLAs to follow negated instructions without additional negation-specific embodied data remains an unexplored problem.

To address these challenges, we propose NegaAlign, a parameter-efficient, plug-and-play framework for following negated instructions without costly negation-specific embodied data for the reliable instruction following. 
Our key insight is that when action expert generates the action sequences, most information are provided by the visual tokens, while the vision tokens are grounded by the instruction text representations.
Specifically, we firstly construct teacher-student instruction pairs from existing affirmative-task data and hand-crafted negated instructions without additional negation-specific action data.
Subsequently, we propose sophisticated Negation Transformation Layers into selected layers of the VLM backbone to revise intermediate instruction representations of negated instructions, which help to modulate the instruction representations to be more compatible with the negated constraints.
Meanwhile, we apply teacher-guided alignment to instruction-relevant visual tokens. 
The alignment propagates the revised instruction representations to the visual tokens which is beneficial for the action generation. Training phase updates only the inserted layers using image--language supervision, while the pretrained backbone and action expert remain frozen.

To evaluate the capability of reliable instruction following, we establish NegaBench, a simulation benchmark spanning 10 scenarios across five domains for systematically evaluating manipulation under negated constraints. 
Experiments on the NegaBench and real-world robot demonstrate that NegaAlign significantly improves the success rate of negated instruction following across various scenarios with fewer update of parameters.
The main contributions of this work are listed as follows:
\begin{itemize}
    \item We present the first investigation of negated instruction understanding and execution in VLAs, revealing how semantics from language instructions are encoded in internal representations and gradually propagate through the model to shape action generation.
    \item We propose NegaAlign, a plug-and-play framework which revises instruction representations through Negation Transformation Layers and propagates negation semantics to visual representations through Teacher-Guided Visual Alignment, enabling VLAs to follow negated constraints efficiently by updating only a few number of parameters.  
    \item We establish NegaBench, a benchmark for systematically evaluating robotic manipulation under negated instructions. Experiments in simulation and real-world robot demonstrate that NegaAlign substantially improves negated instruction following across multiple existing VLAs while preserving performance on affirmative instructions.
\end{itemize}

\section{Related Work}

\subsection{Negation in Language, Vision, and Action}

Negation remains challenging for large language models (LLMs) despite their strong general capabilities. Evaluations show that LLMs degrade on negated prompts, sometimes worsening with scale, and struggle with negated commonsense knowledge \cite{pmlr-v203-jang23a,wang2026commonsense}. Mechanistic analyses further reveal that shortcut behaviors can disrupt how LLMs suppress or reconstruct negated meanings \cite{zhou2026how}. Similar challenges arise in vision-language models (VLMs), where negated semantics must be grounded in visual content. Pretrained VLMs degrade significantly on negated expressions \cite{dobreva2021investigating, singh2024learn}. NeIn~\cite{Bui_2025_CVPR} shows that image-editing models struggle to remove or exclude specified content under negative instructions, while NegBench~\cite{alhamoud2025vision} finds that modern VLMs struggle to distinguish retained from excluded content in negated retrieval and reasoning. Thus, visual negation requires both semantic understanding and correct grounding of exclusion relations.

VLAs such as OpenVLA~\cite{pmlr-v270-kim25c}, $\pi_{0.5}$~\cite{intelligence2025pi05visionlanguageactionmodelopenworld}, and GR00T N series~\cite{bjorck2025gr00t} map language-conditioned visual observations to robot actions. Recent work has begun to address negative and corrective language: Hi Robot~\cite{shi2025hirobotopenendedinstruction} handles negative feedback through hierarchical reasoning, PGPS~\cite{kim2026potential} reduces forbidden-region grasp violations through inference-time steering, and LILAC~\cite{cui2023no} enables online language correction in shared autonomy. However, these methods depend on hierarchical reasoning, external grounding, or online interaction. How negated semantics alter visual grounding within an end-to-end VLA and propagate to action generation remains underexplored.
   
\subsection{Language Conditioning and Instruction Following in VLAs}

Natural language serves as a primary mechanism for task specification and generalization in VLAs \cite{yao2023bridging}. Nevertheless, task success alone does not establish effective linguistic grounding \cite{dong-etal-2026-language}: VLAs may exploit visual shortcuts \cite{zhao2026instructmove}, lose linguistic competence during downstream fine-tuning \cite{zhang2026flatness}, or execute visually plausible actions under semantically contradictory instructions \cite{zhang2026restoring}. Reliable instruction following in VLAs therefore requires the policy to remain sensitive to linguistic semantics, rather than being dominated by visual cues and degenerating into a language-agnostic imitation policy.

Effective language grounding must ultimately propagate from linguistic semantics through visual representations to action generation. Recent work strengthens this coupling by using language and execution history to reweight task-relevant visual tokens \cite{xiao2026ava}. Motivated by this requirement, we propose NegaAlign, a parameter-efficient, plug-and-play framework for following negated instructions without costly negation-specific robot demonstrations.

\section{Method}
\begin{figure*}[h]
\centering
\includegraphics[width=\textwidth]{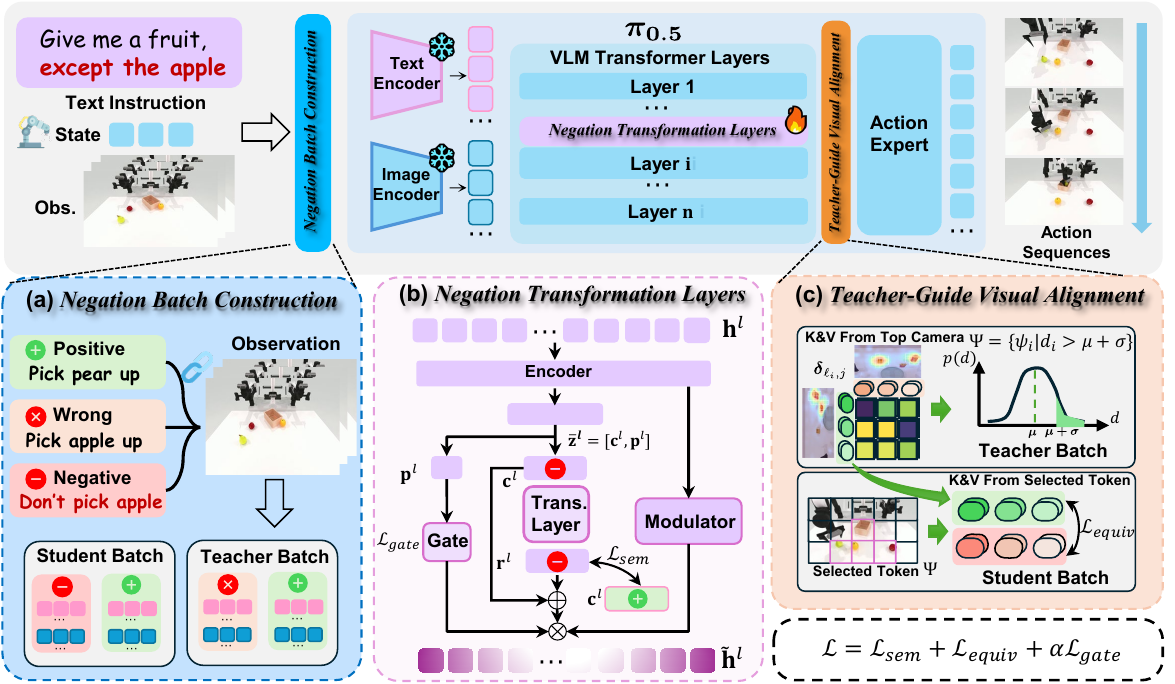}
\caption{Overview of the purposed NegaAlign framework: (a) shows the construction of only vision and language data, (b) Negation Transformation Layer revises intermediate instruction representations, while (c) Teacher-Guided Visual Alignment transfers the revision of language representations to instruction-relevant visual tokens.}
\label{fig:framework}
\end{figure*}
As shown in Figure~\ref{fig:framework}, we present our NegaAlign framework for understanding negation, enabling reliable instruction following. 
We first define the problem in Section~\ref{sec:problem_formulation} and introduce how to construct negation supervision from existing affirmative demonstrations in Section~\ref{sec:data}, then we introduce the Negation Transformation Layer for transferring semantic information of instructions in Section~\ref{sec:ntl} and the Teacher-Guided Visual Alignment for aligning visual representations with negated instructions in Section~\ref{sec:tgva}. Finally, we present the training recipe to make the VLAs adapt to the negated instructions efficiently in Section~\ref{sec:recipe}.

\subsection{Problem Formulation}
\label{sec:problem_formulation}
We assume that a vision--language--action model (VLA) has been pretrained on large-scale robot demonstrations and has acquired the affirmative skills required by the target tasks. 
Our goal is to rapidly adapt it to a given set of target negated instructions and their constraint relations without additional negated action demonstrations.

Following $\pi_{0.5}$~\cite{intelligence2025pi05visionlanguageactionmodelopenworld}, let $f_{\theta}$ denotes the pretrained VLA. Given a visual observation $\mathbf{o}_t$, robot state $\mathbf{s}_t$, and language instruction $\ell$, the model predicts an action sequence of horizon $H$:$\mathbf{a}_{t:t+H-1}= f_{\theta}(\mathbf{o}_t, \mathbf{s}_t, \ell).$
For a negated instruction $\ell^{n}$, successful execution requires both completing the intended task and satisfying its negated constraints. Let $\tau$ denote an execution trajectory, and let $G(\tau,\ell^{n})$ and $C(\tau,\ell^{n})$ indicate task completion and constraint satisfaction, respectively. 
The set of valid trajectories is:
\begin{equation}
    \mathcal{T}_{\mathrm{valid}}(\ell^{n})
    =
    \left\{
        \tau \;\middle|\;
        G(\tau,\ell^{n}) = 1,\;
        C(\tau,\ell^{n}) = 1
    \right\}.
    \label{eq:valid_trajectories}
\end{equation}

We formulate this problem as parameter-efficient post-training using existing affirmative demonstrations. 
Let $\mathbf{x}_0=(\mathbf{o}_0,\mathbf{s}_0)$ denote the initial observation and state, $\mathcal{P}^{n}$ the distribution over initial conditions and target negated instructions, and $p_{\theta,\phi}$ the trajectory distribution induced by the adapted model. 
Its performance is:
\begin{equation}
    J^{n}(\phi)
    =
    \mathbb{E}_{\substack{
        (\mathbf{x}_0,\ell^{n}) \sim \mathcal{P}^{n} \\
        \tau \sim p_{\theta,\phi}(\cdot \mid \mathbf{x}_0,\ell^{n})
    }}
    \left[
        \mathbb{I}
        \left\{
            \tau \in \mathcal{T}_{\mathrm{valid}}(\ell^{n})
        \right\}
    \right].
    \label{eq:negation_success}
\end{equation}
We optimize only lightweight adaptation parameters $\phi$ to maximize $J^{n}(\phi)$ while preserving affirmative instruction-following capability.

\subsection{Negation Supervision Construction}
\label{sec:data}

Existing embodied demonstrations encode individual skills with affirmative instructions in different fine-grained levels, telling robots what to do, such as \textit{pick up the hamburger} or \textit{place the fries on the plate}.
However, people will specify what not to do in daily life, such as \textit{place any food except the chips on the plate}.
Considering the cost of collecting negation-specific robot demonstrations, we construct negation supervision from existing affirmative demonstrations and hand-crafted negated instructions.
Specifically, we expose the latent structure of negation constraint as relational supervision, associating each negated instruction with a set of admissible affirmative interpretations and an explicitly excluded one, as illustrated in Figure~\ref{fig:framework}(a).
The former defines where the negated semantics should be routed, while the latter provides the contrast needed to identify what should be suppressed. 
We formalize this relation as $\mathcal{R}=\left(\ell^{n},\,\mathcal{A},\,\ell^{e}\right)$, 
where $\ell^{n}$ is a negated instruction, $\mathcal{A}=\left\{\ell_{1}^{a},\ldots,\ell_{M}^{a}\right\}$ is the set of affirmative instructions satisfying its constraint, and $\ell^{e}$ describes the excluded behavior.
For example, given \textit{place any foo    d except the chips on the plate}, the instructions for the hamburger and fries belong to $\mathcal{A}$, while the chips instruction is $\ell^{e}$.

During the stage of negation adaption, we first sample a tuple, then one satisfying instruction $\ell^{a}\in\mathcal{A}$, and finally an observation--state pair $(\mathbf{o},\mathbf{s})$ from its affirmative trajectory. 
Then we evaluate four language conditions with keeping the scene fixed:
\begin{equation}
\mathcal{I}_{S}^{a}=(\mathbf{o},\mathbf{s},\ell^{a}), 
\mathcal{I}_{S}^{n}=(\mathbf{o},\mathbf{s},\ell^{n}), 
\mathcal{I}_{T}^{a}=(\mathbf{o},\mathbf{s},\ell^{a}), 
\mathcal{I}_{T}^{e}=(\mathbf{o},\mathbf{s},\ell^{e}),
\label{eq:negation_supervision_tuple}
\end{equation}
where $S$ indicates the student branches and $T$ indicates the teacher branches.
The student branches use $f_{\theta,\phi}$, whereas the teacher branches use the same frozen $f_{\theta}$ without NTLs. 
Their outputs are denoted by $\mathcal{O}_{S}^{a}$, $\mathcal{O}_{S}^{n}$, $\mathcal{O}_{T}^{a}$, and $\mathcal{O}_{T}^{e}$. 
Sharing the observation and state across branches isolates the language contrast and turns existing affirmative data into negation supervision.

\subsection{Negation Transformation Layer}
\label{sec:ntl}
Due to the action generator relies on the attention over visual and language representations, it is difficult for VLAs to learn negation semantics solely from the trajectory that satisfies the instruction.
To this end, we purpose Negation Transformation Layers (NTLs) to revise the intermediate instruction representations of negated instructions to distinguish them from affirmative instructions, thereby enabling the VLA to understand negation semantics.
Specifically, we insert an NTL before each selected VLM layer $l$. The NTL first projects the instruction tokens $\mathbf{h}^l$  into a compact space and pools them into content and polarity embeddings $\mathbf{c}^{l}$ and $\mathbf{p}^{l}$ respectively. 
A lightweight Trans. Layer transforms the $\mathbf{c}^{l}$ into a satisfied affirmative destination $\mathbf{r}^{l}$, while the polarity branch produces a gate $g^{l}=\operatorname{Gate}(\mathbf{p}^l)$, with $a^{l}$ denoting its pre-sigmoid logit used in the gate loss.
To retain token-level distinctions, the resulting semantic shift is modulated by each token representation to obtain $\Delta\mathbf{h}_t^l$:
\begin{equation}
\bar{\mathbf z}^{\,l}
=[\mathbf c^l;\mathbf p^l]
=\operatorname{Pool}\!\left(\operatorname{Encode}(\mathbf h^l)\right),
\mathbf r^l=\operatorname{TransLayer}(\mathbf c^l),
g^l=\operatorname{Gate}(\mathbf p^l),
\label{ntl_1}
\end{equation}
\begin{equation}
\Delta\mathbf{h}_t^l=\operatorname{Modulator}
\!\left(\mathbf r^l-\mathbf c^l;\,\mathbf h_t^l\right),
\qquad
\tilde{\mathbf h}_t^l
=\mathbf h_t^l+
m_tg^l\Delta\mathbf{h}_t^l,
\label{ntl_2}
\end{equation}
where $m_t$ marks instruction tokens. 
Meanwhile, we design an alignment loss $\mathcal{L}^{l}_{\mathrm{sem}}$ to route the revised content toward its paired affirmative representation, and a gate loss $\mathcal{L}^{l}_{\mathrm{gate}}$ to distinguish affirmative from negated inputs:
\begin{equation}
\label{eq:ntl_loss}
\mathcal{L}^{l}_{\mathrm{sem}}=\frac{1}{B}\sum_{b=1}^{B}\left[1-\cos\!\left(\mathbf{r}_{b}^{n,l},
\operatorname{sg}(\mathbf{c}_{b}^{a,l})\right)\right], 
\mathcal{L}^{l}_{\mathrm{gate}}=\frac{1}{2B}\sum_{b=1}^{2B}\operatorname{BCEWithLogits}(a_b^l,y_b),
\end{equation}
where $\mathbf{r}_{b}^{n,l}$ is the negated representation, $\mathbf{c}_{b}^{a,l}$ is the affirmative representation, $a_b^l$ is the pre-sigmoid logit, and $y_b$ is the ground truth.
Since the update is restricted by $m_t$, image, state, and non-instruction prompt tokens remain unchanged. 
NTL thus revises how language selects an acquired skill while preserving the frozen VLA that executes it.

\subsection{Teacher-Guided Visual Alignment}
\label{sec:tgva}
Related work on negation understanding of VLMs \cite{kang2026what} focus on how to enhance the negation semantics in the language space, however, VLAs not only need to understand the negation semantics, but also need to attend to the visual representations that are used for action generation.
Therefore, revising the language representations alone by Negation Transformation Layers~(NTLs) is insufficient to reliably alter the generated actions and the language representation revision should also be propagated to the instruction-relevant visual tokens. 
To achieve this goal, we purpose Teacher-Guided Visual Alignment (TGVA), which guides the revised negation semantics from the language space into the visual representations used for action generation.
After each NTLs, we extract the visual keys and values used by the following attention block and denote their mean cosine distance $d_{b,i}^{l}$ at token $i$ by $D_i^l(\cdot,\cdot)$. 
For each batch sample and layers, TGVA computes:
\begin{equation}
d_{b,i}^{l}=D_i^l(\mathcal{O}_{T,b}^{a},\mathcal{O}_{T,b}^{e}),
\qquad
w_{b,i}^{l}=\frac{\mathbb{I}[i\in\Psi_b^l](d_{b,i}^{l}+\epsilon)}
{\sum_j\mathbb{I}[j\in\Psi_b^l](d_{b,j}^{l}+\epsilon)},
\label{eq:tgva_weights}
\end{equation}
where $\Psi_b^l$ is the set of selected visual-token indices, determined by an adaptive threshold with a top-ranked fallback.
Then, we minimize the weighted distance between the negated student and the satisfying teacher to align the instruction-relevant visual tokens:
\begin{equation}
    \mathcal{L}^{l}_{\mathrm{equiv}}=\frac{1}{B}\sum_{b=1}^{B}\sum_{i\in\Psi_b^l} w_{b,i}^{l}\,D_{i}^{l}\!\left(\mathcal{O}_{S,b}^{n},\operatorname{sg}(\mathcal{O}_{T,b}^{a})\right),
\end{equation}
where $B$ is the batch size.
We apply this alignment to a coordinate-consistent top-camera token and stop gradients through the teacher-derived weights for maintenance of the VLM knowledge. 

\subsection{Training Recipe}
\label{sec:recipe}
Nowadays, the usual training recipe of VLAs is to first pretrain with large-scale embodied demonstration data, then fine-tune it on the task- or scenario-specific demonstration data for adaptation and generalization.
Accordingly, NegaAlign follows a two-stage training recipe after the initial pretraining.
Stage~1 fine-tunes the pretrained VLA on affirmative robot demonstrations, which is consistent with the standard fine-tuning process.
Then during Stage~2, the fine-tuned VLA is frozen, \textbf{Negation Transformation Layers (NTLs)} are inserted, and they are trained with \textbf{Teacher-Guided Visual Alignment (TGVA)} using vision-language data constructed from the existing affirmative demonstrations. 
This stage requires neither negated action trajectories nor additional robot demonstrations.
The training objective of Stage~2 is defined as follows:
\begin{equation}
    \mathcal{L}_{\mathrm{total}}=\frac{1}{|\Omega|}\sum_{l\in\Omega}\left(\mathcal{L}^{l}_{\mathrm{equiv}}+\mathcal{L}^{l}_{\mathrm{sem}}+\alpha\mathcal{L}^{l}_{\mathrm{gate}}\right),
    \label{eq:objectives}
\end{equation}
where $\Omega$ is the set of layers with NTLs, and $\alpha$ is a hyperparameter to balance the three objectives.
The objective is optimized to revise the intermediate instruction representations of negated instructions and propagate the negation semantics to the instruction-relevant visual tokens, thereby enabling the VLA to understand negation semantics and follow negated instructions.

\begin{wrapfigure}{r}{0.5\textwidth}
    \centering
    \includegraphics[width=0.5\textwidth]{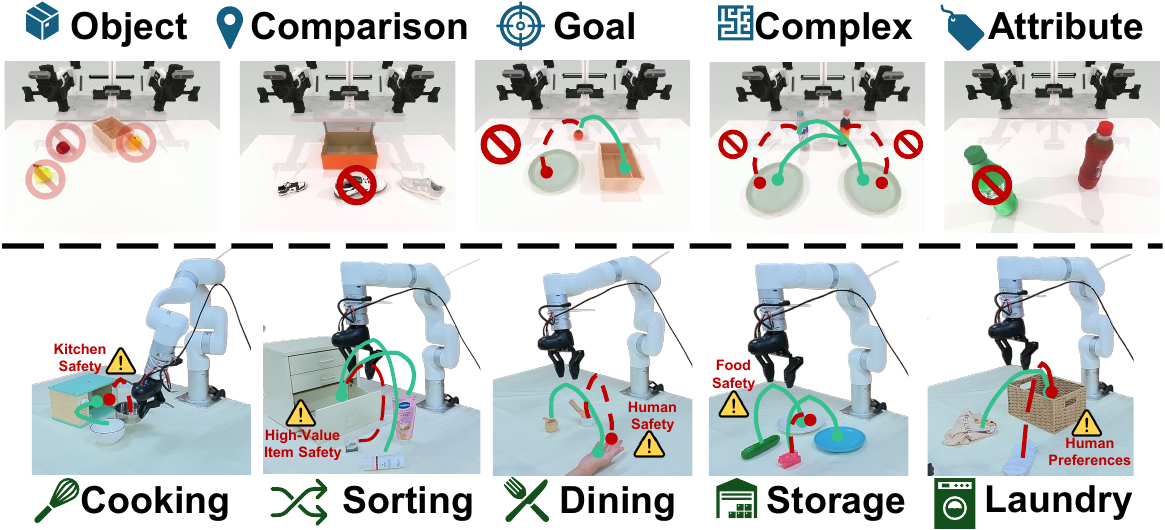}
    \caption{Overview of the NegaBench benchmark. The \textcolor{lightgreen}{\textbf{green trajectory}} indicates a successful execution of the negated instruction, while the \textcolor{darkred}{\textbf{red trajectory}} indicates that the robot failed to meet the negated constraints.}
    \label{fig:benchmark}
\end{wrapfigure}
\subsection{Negation Instruction Following Benchmark Construction}
\label{sec:negabench}

To systematically evaluate whether robots can follow negated constraints during manipulation, we construct NegaBench, a controlled and fine-grained benchmark built upon the RoboTwin simulation environment~\cite{chen2026robotwin}. 
NegaBench comprises 10 manipulation scenarios spanning five representative categories of negation: object exclusion, attribute exclusion, relative comparison, goal-location exclusion, and combinations of multiple constraints. 
These categories progress from selecting among individual objects to executing complex behaviors that must satisfy several exclusions simultaneously. 
Each scenario provides semantically paired affirmative and negated instructions together with explicit constraint-aware success criteria: an execution is considered successful only if the robot completes the shared manipulation goal without violating any excluded object, attribute, or outcome specified by the instruction. 
By evaluating paired instructions under the same scene, manipulation skills, and success conditions, NegaBench controls for basic task competence and scene difficulty, directly measures whether a negated constraint changes the behavioral choice of robot, and systematically exposes the limitations of existing VLA models in constraint-aware instruction following. 
Complete benchmark construction and instruction relations are provided in Appendix~\ref{sec:data_pipeline}, and the evaluation protocol is detailed in Appendix~\ref{sec:simulation_eval_protocol}.

\section{Experiments}
\subsection{Experimental Setup and Evaluation Metrics}

We evaluate NegaAlign on three VLA models, GR00T, $\pi_{0}$, and $\pi_{0.5}$, under three training settings:
\textit{Pos.}, fine-tune the pretrained model with only affirmative instructions; 
\textit{Pos.+Neg.}, joint training on both affirmative and negated trajectories to enable the capability of negation understanding and prevent the catastrophic forgetting of affirmative instructions; 
and \textit{Pos.+Ours}, which applies NegaAlign to an affirmative-trained model. 
We additionally evaluate APT~\cite{xu2026aptactionexpertpretraining} and Qwen-$\pi_{0.5}$ as baselines. Qwen-$\pi_{0.5}$ uses Qwen3-VL-2B-Instruct \cite{bai2025qwen3vltechnicalreport} to rewrite negated instruction base on top camera observation to an affirmative command for an affirmative-trained $\pi_{0.5}$ policy.
Further implementation details for these baselines are given in Appendix~\ref{sec:backbone_details}.
The training detail is provided in Appendix~\ref{sec:training_details}.
We report success rates separately for affirmative instructions (Pos. SR) and negated instructions (Neg. SR). 
An episode is considered successful only if the robot completes the intended manipulation while satisfying all negated constraints.

\begin{table*}[t]
\caption{Comparison of our method with baselines on NegaBench. The best results are highlighted in \textbf{bold}.}
\label{tab:Comparison}
\centering
\setlength{\tabcolsep}{3.1pt}
\renewcommand{\arraystretch}{1.12}
\resizebox{\textwidth}{!}{%
\begin{tabular}{cccccccccccccccc}
\toprule
\multirow{2}{*}{Model} & \multicolumn{1}{c}{\multirow{2}{*}{Training}} & \multicolumn{2}{c}{Object} & \multicolumn{2}{c}{Attribute} & \multicolumn{2}{c}{Comparison} & \multicolumn{2}{c}{Goal} & \multicolumn{2}{c}{Complex} & \multicolumn{2}{c}{Avg.} & \multirow{2}{*}{\shortstack{Trainable\\Param.}} & \multirow{2}{*}{\shortstack{Fine-tuning\\Time(hours)}} \\
\cmidrule(lr){3-4}\cmidrule(lr){5-6}\cmidrule(lr){7-8}\cmidrule(lr){9-10}\cmidrule(lr){11-12}\cmidrule(lr){13-14}
 & & Pos. & Neg. & Pos. & Neg. & Pos. & Neg. & Pos. & Neg. & Pos. & Neg. & Pos. & Neg. & & \\
\midrule
GR00T & Pos. FT & 60.09 & 1.25 & 63.75 & 0.00 & 70.00 & 4.00 & 75.00 & 0.00 & 86.63 & 0.00 & 71.09 & 1.05 & -- & -- \\
$\pi_0$ & Pos. FT & 30.75 & 14.75 & 58.50 & 44.00 & 75.50 & 18.00 & 96.25 & 0.00 & 80.00 & 0.25 & 68.20 & 15.40 & -- & -- \\
$\pi_{0.5}$ & Pos. FT & 91.59 & 3.50 & 88.00 & 0.50 & 89.00 & 7.00 & 96.00 & 0.00 & 99.25 & 2.00 & 92.77 & 2.60 & -- & -- \\
APT & Pos. FT & 58.92 & 6.00 & 75.50 & 13.50 & 73.00 & 5.00 & 84.75 & 13.50 & 85.13 & 3.00 & 75.46 & 8.20 & -- & -- \\
Qwen-$\pi_{0.5}$ & Pos. FT & 49.50 & 52.00 & 84.25 & 94.00 & 56.75 & 25.50 & 93.50 & 60.50 & 73.88 & 46.50 & 71.58 & 55.70 & -- & -- \\
\midrule
\midrule
\multirow{2}{*}{GR00T}
 & \cellcolor{red!7}Pos. \& Neg. FT & \cellcolor{red!7}60.92 & \cellcolor{red!7}44.00 & \cellcolor{red!7}60.00 & \cellcolor{red!7}65.50 & \cellcolor{red!7}71.50 & \cellcolor{red!7}65.00 & \cellcolor{red!7}91.25 & \cellcolor{red!7}77.50 & \cellcolor{red!7}93.63 & \cellcolor{red!7}92.25 & \cellcolor{red!7}75.46 & \cellcolor{red!7}68.85 & \cellcolor{red!7}567.6M/3.2B  & \cellcolor{red!7}87.6 \\
 & \cellcolor{green!9}\textbf{w/ Ours} & \cellcolor{green!9}59.84 & \cellcolor{green!9}24.75 & \cellcolor{green!9}68.75 & \cellcolor{green!9}68.00 & \cellcolor{green!9}72.75 & \cellcolor{green!9}61.50 & \cellcolor{green!9}76.00 & \cellcolor{green!9}87.50 & \cellcolor{green!9}87.13 & \cellcolor{green!9}47.00 & \cellcolor{green!9}72.89 & \cellcolor{green!9}\textbf{57.75} & \cellcolor{green!9}\textbf{4.35M/3.2B} & \cellcolor{green!9}\textbf{2.7} \\
\midrule
\multirow{2}{*}{$\pi_0$}
 & \cellcolor{red!7}Pos. \& Neg. FT & \cellcolor{red!7}29.92 & \cellcolor{red!7}26.75 & \cellcolor{red!7}74.75 & \cellcolor{red!7}83.00 & \cellcolor{red!7}77.75 & \cellcolor{red!7}72.50 & \cellcolor{red!7}98.50 & \cellcolor{red!7}94.00 & \cellcolor{red!7}77.50 & \cellcolor{red!7}80.00 & \cellcolor{red!7}71.68 & \cellcolor{red!7}71.25 & \cellcolor{red!7}468.0M/3.6B & \cellcolor{red!7}251.4 \\
 & \cellcolor{green!9}\textbf{w/ Ours} & \cellcolor{green!9}29.42 & \cellcolor{green!9}24.75 & \cellcolor{green!9}59.50 & \cellcolor{green!9}59.50 & \cellcolor{green!9}76.50 & \cellcolor{green!9}59.50 & \cellcolor{green!9}96.00 & \cellcolor{green!9}96.50 & \cellcolor{green!9}80.63 & \cellcolor{green!9}76.25 & \cellcolor{green!9}68.41 & \cellcolor{green!9}\textbf{63.30} & \cellcolor{green!9}\textbf{11.6M/3.6B} & \cellcolor{green!9}\textbf{3.3} \\
\midrule
\multirow{2}{*}{$\pi_{0.5}$}
 & \cellcolor{red!7}Pos. \& Neg. FT & \cellcolor{red!7}85.67 & \cellcolor{red!7}77.75 & \cellcolor{red!7}90.50 & \cellcolor{red!7}93.00 & \cellcolor{red!7}85.75 & \cellcolor{red!7}81.00 & \cellcolor{red!7}96.75 & \cellcolor{red!7}92.50 & \cellcolor{red!7}98.38 & \cellcolor{red!7}98.75 & \cellcolor{red!7}91.41 & \cellcolor{red!7}88.60 & \cellcolor{red!7}467.0M/3.7B & \cellcolor{red!7}179.9 \\
 & \cellcolor{green!9}\textbf{w/ Ours} & \cellcolor{green!9}87.67 & \cellcolor{green!9}77.50 & \cellcolor{green!9}91.75 & \cellcolor{green!9}94.50 & \cellcolor{green!9}89.50 & \cellcolor{green!9}78.00 & \cellcolor{green!9}94.25 & \cellcolor{green!9}93.50 & \cellcolor{green!9}99.38 & \cellcolor{green!9}98.75 & \cellcolor{green!9}92.51 & \cellcolor{green!9}\textbf{88.45} & \cellcolor{green!9}\textbf{11.6M/3.7B} & \cellcolor{green!9}\textbf{4.6} \\
\bottomrule
\end{tabular}}
\end{table*}

\begin{figure*}[htbp]
\centering
\includegraphics[width=\textwidth]{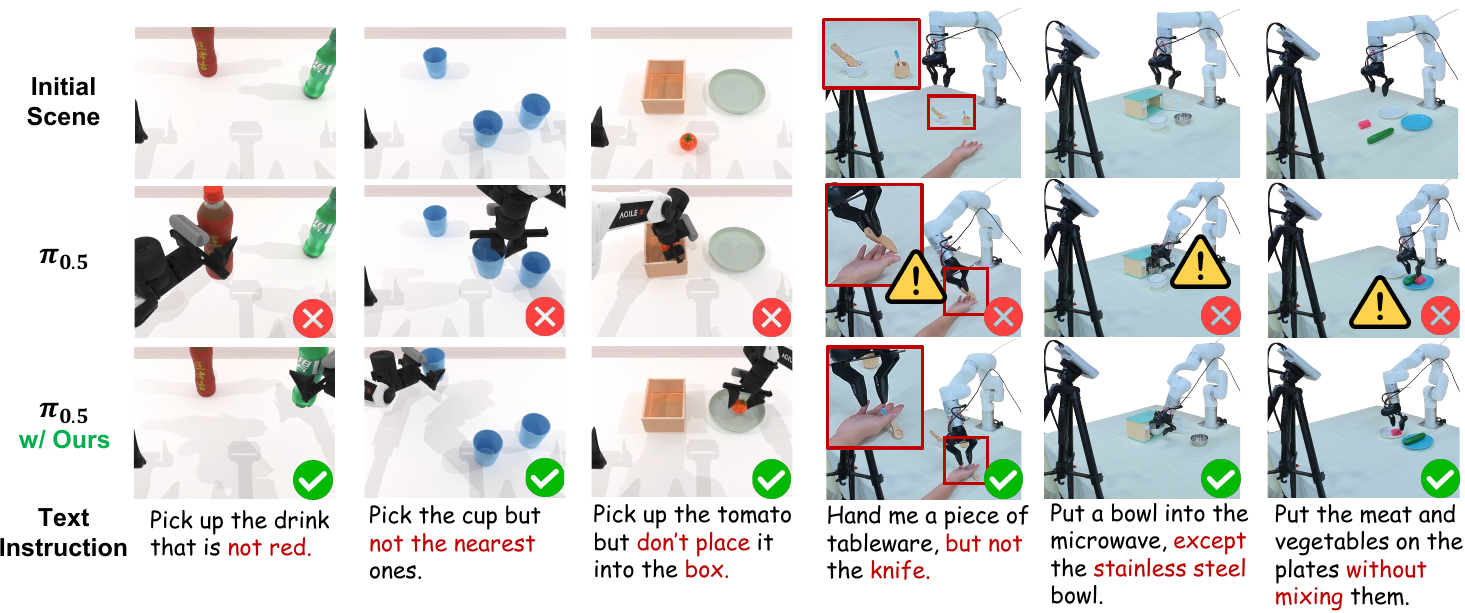}
\caption{The visualization of the simulation and real-world experiments. The robot is required to follow negated instructions in various scenarios. }
\label{fig:visualization}
\end{figure*}
\subsection{Comparison Results on NegaBench}
As shown in Table~\ref{tab:Comparison}, affirmative-only training yields strong affirmative performance but generalizes poorly to negated instructions: GR00T, $\pi_0$, and $\pi_{0.5}$ achieve average negated success rates of 1.05\%, 15.40\%, and 2.60\%, respectively. 
Explicit negated-trajectory training raises these rates to 68.85\%, 71.25\%, and 88.60\%. 
This contrast motivates NegaAlign to ground negated constraints in existing manipulation skills rather than relearn the skills themselves.

APT reaches 75.46\% affirmative success but only 8.20\% negated success with affirmative-only training, indicating that preserving the capabilities of VLM in a VLA is insufficient to solve the negated instruction follow problem. 
Qwen-$\pi_{0.5}$ directly rewrites negated instructions into affirmative commands, yet achieves only 55.70\% negated success and 71.58\% affirmative success. 
This approach adds an upstream VLM inference step at every action-chunk request, increasing inference latency and making the action depend on the VLM visual grounding and instruction rewrite capabilities. 
Its substantially lower negated success than 88.45\% of our purposed NegaAlign suggests that direct instruction rewriting is not an effective solution to negated instruction following in this setting.

Without using any negated action trajectories, NegaAlign reaches 57.75\%, 63.30\%, and 88.45\% negated success on GR00T, $\pi_0$, and $\pi_{0.5}$, gains of 56.70, 47.90, and 85.85 percentage points over their corresponding \textit{Pos.} baselines. Affirmative success is 72.89\%, 68.41\%, and 92.51\%, respectively, compared with 71.09\%, 68.20\%, and 92.77\% for \textit{Pos. FT}.
Averaged across the three models, it achieves 69.83\% negated success, reaching 91.6\% of the performance obtained through direct negated-trajectory training. 
On $\pi_{0.5}$, NegaAlign achieves 88.45\%, nearly matching the 88.60\% of \textit{Pos.\& Neg.} 
These results demonstrate that semantic supervision constructed from existing affirmative data enables effective negated instruction following without additional robot demonstrations.

\subsection{Ablation Studies and Mechanism Analysis}
\label{sec:ablation}

\begin{figure*}[htbp]
    \centering
    \includegraphics[width=\linewidth]{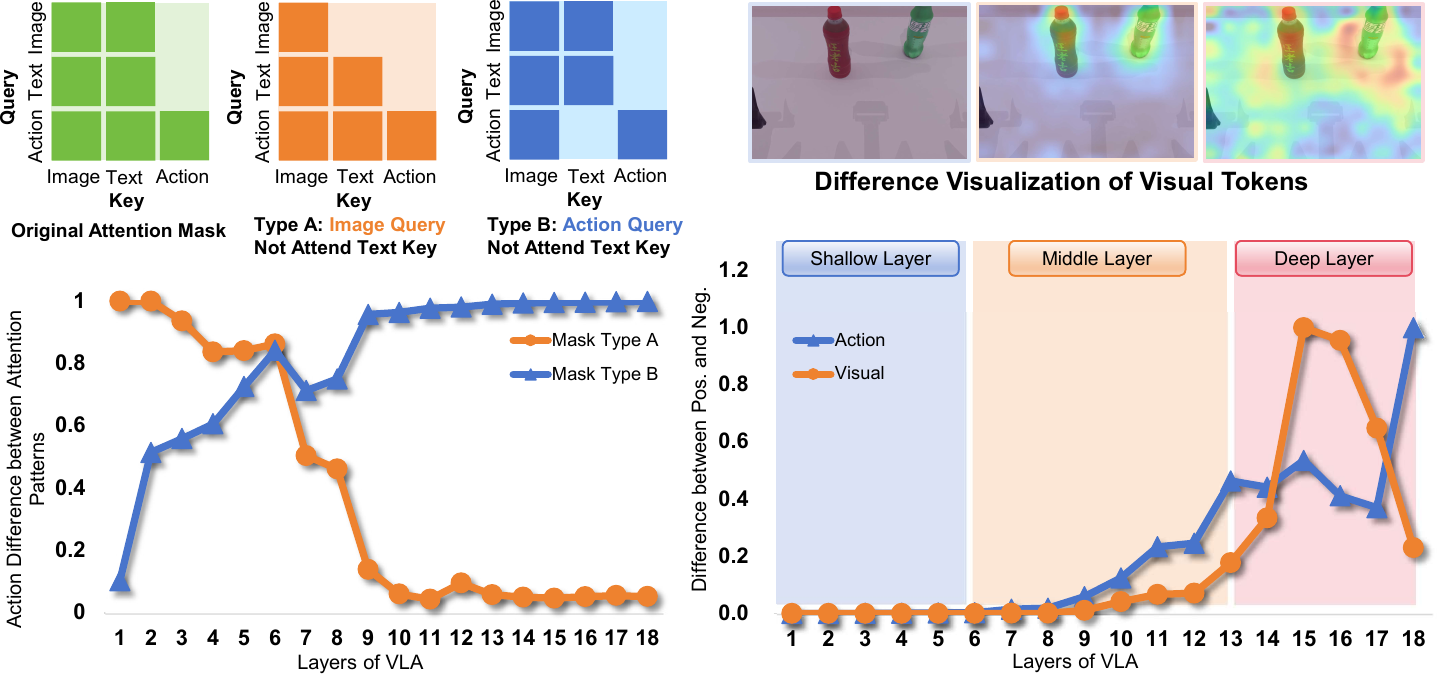}
    \caption{Left: Normalized instruction-induced action separation under attention masking.  Right: Layer-wise differences between satisfying and excluded instructions.}
    \label{fig:analysis}
\end{figure*}

\begin{wraptable}{r}{0.5\textwidth}
\vspace{-0.8\baselineskip}
\caption{Ablation study on the effect of TGVA and different layer configurations of NTL. }
\label{tab:ablation}
\centering
\footnotesize
\renewcommand{\arraystretch}{1.40}
\setlength{\tabcolsep}{2.6pt}
\resizebox{\linewidth}{!}{%
\begin{tabular}{cccccc}
\toprule
\multirow{2}{*}{Setting} & \multirow{2}{*}{Layer} & \multirow{2}{*}{TGVA} & \multirow{2}{*}{NTL} & \multicolumn{2}{c}{Success Rate (\%)} \\
\cmidrule(lr){5-6}
& & & & Pos. & Neg. \\
\midrule
Baseline & -- & \textcolor{red!70!black}{\XSolidBrush} & \textcolor{red!70!black}{\XSolidBrush} & \makebox[5.7em][c]{92.77\rlap{\hspace{0.15em}{\fontsize{5.5}{6}\selectfont\textcolor{green!65!black}{(+0.26)}}}} & \makebox[5.7em][c]{2.60\rlap{\hspace{0.15em}{\fontsize{5.5}{6}\selectfont\textcolor{red!65!black}{(-85.85)}}}} \\
Exp. 1 & \cellcolor[HTML]{C9DAF8}Shallow & \textcolor{green!45!black}{\CheckmarkBold} & \textcolor{green!45!black}{\CheckmarkBold} & \makebox[5.7em][c]{92.26\rlap{\hspace{0.15em}{\fontsize{5.5}{6}\selectfont\textcolor{red!65!black}{(-0.25)}}}} & \makebox[5.7em][c]{85.10\rlap{\hspace{0.15em}{\fontsize{5.5}{6}\selectfont\textcolor{red!65!black}{(-3.35)}}}} \\
Exp. 2 & \cellcolor[HTML]{C9DAF8}Deep & \textcolor{green!45!black}{\CheckmarkBold} & \textcolor{green!45!black}{\CheckmarkBold} & \makebox[5.7em][c]{91.92\rlap{\hspace{0.15em}{\fontsize{5.5}{6}\selectfont\textcolor{red!65!black}{(-0.59)}}}} & \makebox[5.7em][c]{62.20\rlap{\hspace{0.15em}{\fontsize{5.5}{6}\selectfont\textcolor{red!65!black}{(-26.25)}}}} \\
Exp. 3 & Middle & \cellcolor[HTML]{F4CCCC}\textcolor{red!70!black}{\XSolidBrush} & \textcolor{green!45!black}{\CheckmarkBold} & \makebox[5.7em][c]{91.27\rlap{\hspace{0.15em}{\fontsize{5.5}{6}\selectfont\textcolor{red!65!black}{(-1.24)}}}} & \makebox[5.7em][c]{21.35\rlap{\hspace{0.15em}{\fontsize{5.5}{6}\selectfont\textcolor{red!65!black}{(-67.10)}}}} \\
Exp. 4 & Middle & \textcolor{green!45!black}{\CheckmarkBold} & \cellcolor[HTML]{FCE5CD}\textcolor{red!70!black}{\XSolidBrush} & \makebox[5.7em][c]{40.52\rlap{\hspace{0.15em}{\fontsize{5.5}{6}\selectfont\textcolor{red!65!black}{(-51.99)}}}} & \makebox[5.7em][c]{86.65\rlap{\hspace{0.15em}{\fontsize{5.5}{6}\selectfont\textcolor{red!65!black}{(-1.80)}}}} \\
\rowcolor[HTML]{D9EAD3}
Ours & Middle & \textcolor{green!45!black}{\CheckmarkBold} & \textcolor{green!45!black}{\CheckmarkBold} & \makebox[5.7em][c]{\textbf{92.51}} & \makebox[5.7em][c]{\textbf{88.45}} \\
\bottomrule
\end{tabular}}
\vspace{-0.5\baselineskip}
\end{wraptable}

\paragraph{How do language semantics influence action generation?}
We first examine how language information affects the action generator. 
Figure~\ref{fig:analysis} in left reports the action separation between satisfying and excluded instructions under two attention masks, normalized by the separation under full attention, where $1$ indicates full preservation. 
When action queries cannot attend to text keys, the separation remains near 1 in deeper layers, showing that direct language access has little effect on the final actions. 
In contrast, blocking image queries from text keys causes the separation to approach zero, leaving the policy unable to distinguish the two instructions at the action level. 
Language semantics therefore primarily condition visual representations, which subsequently guide the action expert to generate actions. 
This observation of visual mediation motivates the design of Teacher-Guide Visual Alignment (TGVA).

\paragraph{Why is Teacher-Guide Visual Alignment(TGVA) effective?}
\label{TGVA_effective}
Based on the above analysis, we visualize the layer-wise differences of visual tokens between satisfying and excluded instructions in Figure~\ref{fig:analysis} right.
The visual differences between instructions are negligible in shallow layers, emerge and concentrate on instruction-relevant objects in middle layers, and spread toward background and irrelevant regions in deeper layers. 
Therefore, TGVA is designed to align the negated branch with the satisfying teacher in middle layers, where semantic visual grounding has formed but has not yet become diffuse. 
Removing TGVA reduces negated success from 88.45\% to 21.35\%, while affirmative success remains relatively high at 91.27\%, as Table~\ref{tab:ablation} shows.  
It indicates that NTL could preserve the original manipulation capability but cannot reliably propagate the revised semantics into action generation without TGVA, confirming the necessity of visual alignment and the effectiveness of the visual mediation mechanism identified in Figure~\ref{fig:analysis} left.

\paragraph{What is the optimal intervention depth for Negation Transformation Layers(NTL)?}

As shown in Table~\ref{tab:ablation}, inserting NTLs into middle layers of the VLM backbone yields the best performance in both affirmative and negated instruction following. 
Meanwhile, moving NTL to shallow or deep layers reduces negated success to 85.10\% and 62.20\%, respectively, while affirmative performance remains close to 92\%.
It indicates that the instruction-dependent visual representations have different characteristics in different layers, and the middle layers provide the best intervention point for NTLs, after semantic grounding emerges but before it becomes diffuse, which is consistent with the analysis in Figure~\ref{fig:analysis} right.

\paragraph{Why is the NTL design necessary?}
As shown in Table~\ref{tab:ablation}, retaining TGVA while removing NTL yields 86.65\% negated success but collapses affirmative performance from 92.51\% to 40.52\%.
In this variant, the proposed NTL is replaced by a trainable ungated revision layer at the same insertion positions, while all pretrained VLA parameters remain frozen.
The replacement applies revision to both affirmative and negated instructions and is optimized using the semantic alignment and TGVA objectives.
The decrease in performance highlights the importance of NTL in maintaining the performance of affirmative instructions.
Specifically, visual alignment can induce constraint-satisfying behavior, but without controlled semantic transformation it also disrupts affirmative instructions.
NTL separates semantic content from polarity and gates revision only when needed, enabling the complete model to achieve 88.45\% negated success while retaining 92.51\% affirmative success.
Thus, NTL prevents this revision from compromising the original capabilities of policy while TGVA propagates revised semantics through action-relevant visual tokens.

\subsection{Real-World Robot Experiments}
\begin{wraptable}{r}{0.5\textwidth}
\vspace{-0.8\baselineskip}
\caption{Real-world success counts on five xArm manipulation tasks. The best results are highlighted in \textbf{bold}.}
\label{tab:real-world}
\centering
\footnotesize
\renewcommand{\arraystretch}{1}
\setlength{\tabcolsep}{3.2pt}
\resizebox{\linewidth}{!}{%
\begin{tabular}{c
  >{\columncolor{baseCell}}c >{\columncolor{baseCell}}c
  >{\columncolor{oursCell}}c >{\columncolor{oursCell}}c}
\toprule
\multirow{2}{*}{Task}
  & \multicolumn{2}{c}{$\pi_{0.5}$}
  & \multicolumn{2}{c}{$\pi_{0.5}$ w/ Ours} \\
\cmidrule(lr){2-3}\cmidrule(lr){4-5}
& \cellcolor{white}Pos. & \cellcolor{white}Neg. & \cellcolor{white}Pos. & \cellcolor{white}Neg. \\
\midrule
Sorting       & 46/50 &  9/50 & 42/50 & \textbf{40/50} \\
Laundry         & 46/50 &  4/50 & 49/50 & \textbf{47/50} \\
Storage       & 45/50 &  1/50 & 44/50 & \textbf{50/50} \\
Cooking    & 49/50 &  0/50 & 48/50 & \textbf{48/50} \\
Dining         & 43/50 & 17/50 & 44/50 & \textbf{37/50} \\
\midrule
Avg. SR (\%)        & 91.6 & 12.4 & 90.8 & \textbf{88.8} \\
Avg. Inference Latency (ms)
  & \multicolumn{2}{>{\columncolor{baseCell}}c}{\textbf{481.6}}
  & \multicolumn{2}{>{\columncolor{oursCell}}c}{494.3} \\
\bottomrule
\end{tabular}}

\end{wraptable}

We further evaluate NegaAlign on a physical xArm robot in five safety-critical household scenarios, as illustrated in Figure~\ref{fig:benchmark}.
A detailed description of the real-world task setting is provided in Appendix~\ref{sec:realworld_eval_details}.
Violating these negated constraints can cause property loss or damage, and potentially serious injury, making correct exclusion-aware behavior essential. 
Each negated instruction is evaluated over 50 trials. 
We compare the affirmative-only $\pi_{0.5}$ policy with the same policy adapted by NegaAlign, as shown in Table~\ref{tab:real-world}.

The affirmative-only policy achieves a high affirmative success rate of 91.6\%, but succeeds in only 12.4\% of negated trials. 
This large discrepancy shows that robust manipulation alone does not enable a VLA to reliably respect exclusion constraints.
With our method, the average negated success rate rises to 88.8\%, a gain of 76.4 percentage points, while affirmative performance remains essentially unchanged at 90.8\%. 
The consistent improvement across diverse household scenarios demonstrates that our approach effectively transfers negation understanding into real-world robot behavior without compromising the original manipulation capabilities of policy.

\section{Conclusion}
We present NegaAlign, a parameter-efficient framework that combines Negation Transformation Layers and Teacher-Guided Visual Alignment to enable negated instruction following in pretrained VLAs without additional negated action trajectories. We also introduce NegaBench to evaluate this capability in robotic manipulation. Simulation and real-world experiments demonstrate substantial improvements in negated instruction following while preserving affirmative performance.

The limitations of this work include the reliance on existing pretrained VLAs, which may not cover all possible negated scenarios, and the potential for performance degradation in highly complex or ambiguous negated instructions that leads our direction of future work. 

\section*{AI Use Statement}
In this work, we used generative AI tools to assist with language editing, figure presentation, checking mathematical notation, and checking the consistency between the manuscript and existing code. We did not use generative AI to generate research data, formulate or prove mathematical claims, design experiments, or analyze or interpret experimental results. The authors reviewed all AI-assisted text and suggestions and verified their accuracy against the original code, experimental records, and relevant literature. The authors take full responsibility for the final text, claims, and research artifacts in this work.

\bibliography{iclr2027_conference}
\bibliographystyle{iclr2027_conference}
\clearpage
\appendix
\raggedbottom
\section{Implementation and Training Details}
\label{sec:training_details}

\subsection{Backbones, Trainable Modules, and Branch Execution}
\label{sec:backbone_details}
We instantiate NegaAlign from an affirmative-only checkpoint of each VLA. During negation post-training, all parameters inherited from the checkpoint are frozen, including the visual encoder, the vision--language backbone, the action expert or action head, and any adapters learned during affirmative training. The only trainable parameters are the inserted Negation Transformation Layers (NTLs). For $\pi_0$ and $\pi_{0.5}$, we insert NTLs before VLM layers 6--13 (one-based indexing). For GR00T, whose vision--language interface contains four Vision--Language Self--Attention (VLSA) blocks, we insert NTLs before the first three blocks and use the following block to obtain the Key and Value supervision signal. These configurations introduce 11.6M trainable parameters for $\pi_0$ and $\pi_{0.5}$ and 4.35M for GR00T.

The additional APT baseline uses its affirmative-trained checkpoint and executes dual-arm end-effector pose and gripper actions. Qwen-$\pi_{0.5}$ pairs the affirmative-trained $\pi_{0.5}$ checkpoint with an unmodified Qwen3-VL-2B-Instruct model. At each action-chunk request, Qwen receives the current top-camera image and original instruction and generates a command for $\pi_{0.5}$; both affirmative and negated instructions pass through this rewriting step. Table~\ref{tab:qwen-rewrites} shows examples of the generated commands.

\begin{table*}[t]
\caption{Examples of Qwen-$\pi_{0.5}$ instruction rewrites on NegaBench. The generated commands are shown as produced, including cases where a negated constraint is retained or omitted.}
\label{tab:qwen-rewrites}
\centering
\footnotesize
\setlength{\tabcolsep}{4pt}
\renewcommand{\arraystretch}{1.12}
\begin{tabular}{@{}p{0.14\textwidth}p{0.395\textwidth}p{0.395\textwidth}@{}}
\toprule
Type & Original instruction & Qwen-generated command \\
\midrule
Attribute & pick up the drink that is not red & pick up the green drink \\
 & pick the cup that is not full & pick the blue cup \\
\addlinespace[2pt]
Comparison & pick the cup but not the nearest one & pick the middle cup \\
 & put one shoe in the shoebox, don't choose the biggest one & put the white shoe in the shoebox \\
\addlinespace[2pt]
Complex & pick up the bottle but not the water and place it on the plate but not the left one & pick up the bottle and place it on the right plate \\
 & pick up the bottle but not the soy sauce and place it on the plate but not the right one & pick up the water bottle and place it on the left plate \\
 & pick up a dumbbell, not the red one and not the biggest one & pick up a dumbbell, not the red one and not the biggest one \\
 & pick up a dumbbell, not the black one and not the smallest one & pick up a red dumbbell \\
\addlinespace[2pt]
Goal & pick up the tomato but don't place it into the box & pick up the tomato and place it on the table \\
 & pick up the bread but don't place it on the table & pick up the bread and place it on the tray \\
\addlinespace[2pt]
Object & pick up one fruit but not apple into the box & pick up the orange into the box \\
 & place any food except the chips on the plate & place any food on the plate \\
 & place any food except the fries on the plate & place any food on the plate \\
\bottomrule
\end{tabular}
\end{table*}

The configured batch size $B$ counts independently sampled affirmative scene--instruction pairs. Each pair is expanded into four logical language-conditioned branches over the same observation and robot state: the affirmative student $S^a$, negated student $S^n$, satisfying teacher $T^a$, and excluded teacher $T^e$. Consequently, a training step with $B=4$ evaluates 16 branch instances, but this should not be interpreted as either four model replicas or an optimization batch size of 16. All branches share one set of VLA weights. The two teachers are stateless, no-gradient forward branches of the same frozen checkpoint with the NTLs bypassed; they are neither separately trained networks nor exponential-moving-average copies. The number of independent scenes contributing to the stochastic gradient remains four.

For $\pi_0$ and $\pi_{0.5}$, the two student branches are concatenated and executed as one gradient-carrying batch of size $2B$. The satisfying and excluded teachers are then evaluated as two sequential no-gradient batches of size $B$. Thus, when $B=4$, the largest forward that retains activations for backpropagation contains eight samples, while each teacher forward contains four. Frozen image encoding and prompt embedding are also evaluated without constructing an autograd graph. This scheduling provides the four required representations without storing the activations of all $4B$ branches simultaneously.

GR00T further reuses computation across branches. Its frozen Qwen backbone jointly processes the satisfying, negated, and excluded language variants as a batch of size $3B$; the satisfying hidden states are shared by $T^a$ and $S^a$. The subsequent VLSA computation consists of two no-gradient teacher branches of size $B$ and one gradient-carrying student branch of size $2B$. Images are preprocessed once per sampled scene and repeated only after preprocessing. Hence, with $B=4$, Qwen receives 12 language variants, while the downstream VLSA blocks realize the same 16 logical branch instances required by the objective.

Post-training executes only the visual--language prefix and the selected VLM blocks required to construct the intermediate supervision. It does not execute the action expert and does not consume action targets. This substantially reduces memory relative to running four complete VLAs, although compute still scales with the teacher and student branch forwards described above. At inference time, all teacher branches are discarded. The input instruction is revised by the learned NTLs at the same insertion sites, after which the frozen action-generation procedure of the original VLA is executed unchanged.

\subsection{NTL Architecture}
Each selected VLM layer has an independently parameterized NTL; parameters are not shared across insertion sites. Let $\mathbf{h}_t^l\in\mathbb{R}^{d}$ be hidden state of token $t$ at site $l$, and let $m_t\in\{0,1\}$ indicate whether it belongs to the natural-language instruction. The tokenizer-derived mask excludes image and state tokens, special prompt tokens, and padding. The NTL first normalizes and compresses the selected tokens to a 256-dimensional bottleneck, then pools them into an instruction-level representation:
\begin{equation}
\mathbf{z}_t^l=W_{\mathrm{down}}^l\operatorname{LN}(\mathbf{h}_t^l),
\qquad
\bar{\mathbf{z}}^l=
\frac{\sum_t m_t\mathbf{z}_t^l}{\max(1,\sum_t m_t)},
\qquad
[\mathbf{c}^l;\mathbf{p}^l]=\bar{\mathbf{z}}^l,
\label{eq:appendix-ntl-encode}
\end{equation}
where $\mathbf{c}^l\in\mathbb{R}^{224}$ is the content code and $\mathbf{p}^l\in\mathbb{R}^{32}$ is the polarity code. The split reserves most of the bottleneck for task content while exposing a compact subspace from which the module can decide whether semantic revision is required.

The transformation layer is a $224\!\rightarrow\!896\!\rightarrow\!224$ MLP, while a separate linear predictor maps the polarity code to a scalar gate:
\begin{equation}
\mathbf{r}^l=W_2^l\operatorname{GELU}(W_1^l\mathbf{c}^l),
\qquad
a^l=W_g^l\mathbf{p}^l,
\qquad
g^l=\sigma(a^l).
\label{eq:appendix-ntl-resolve}
\end{equation}
Here, $\mathbf{r}^l$ is the satisfying affirmative destination predicted from the instruction content. A single global displacement $\mathbf{r}^l-\mathbf{c}^l$ would update all words identically, so we additionally compute a token-specific modulation vector $\mathbf{u}_t^l=\tanh(W_m^l\mathbf{z}_t^l)\in\mathbb{R}^{224}$. The complete update is
\begin{equation}
\Delta\mathbf{h}_t^l
=W_{\mathrm{up}}^l\!\left(
\mathbf{u}_t^l\odot(\mathbf{r}^l-\mathbf{c}^l)
\right),
\qquad
\tilde{\mathbf{h}}_t^{l}
=\mathbf{h}_t^l+m_tg^l\Delta\mathbf{h}_t^l.
\label{eq:appendix-ntl-update}
\end{equation}
The $224\!\rightarrow\!d$ up-projection returns the residual to the host-model width. The token modulator allows object, attribute, relation, and action words to respond differently to the same instruction-level revision. The mask scatters the revised states back only to their original instruction positions; all other tokens remain bitwise unchanged at the NTL output. Gate supervision drives $g^l$ toward zero for affirmative inputs and toward one for negated inputs, so the same module can preserve the original affirmative pathway while activating the learned revision when needed. Each NTL contributes approximately 1.45M parameters for the backbones used in our experiments, yielding 11.6M parameters over eight sites for $\pi_0$ and $\pi_{0.5}$ and 4.35M over three sites for GR00T.

\subsection{Teacher-Guided Visual Alignment}
TGVA uses two frozen forward branches to determine both \emph{where} negation changes visual grounding and \emph{which} affirmative grounding the negated student should approach. For every sampled scene, the satisfying teacher $\mathcal{O}_{T}^{a}$ receives the selected admissible affirmative instruction and the excluded teacher $\mathcal{O}_{T}^{e}$ receives the prohibited affirmative instruction. They share exactly the same image and robot state. Both branches bypass all NTLs and run without gradients, ensuring that their difference is caused only by the language condition rather than scene variation or teacher adaptation.

An NTL is applied immediately before selected layer $l$. We execute that layer and observe the visual keys and values projected at the input of layer $l+1$; hence the observer measures the first visual representation that has incorporated the revised language through a complete attention block. For $\pi_0$ and $\pi_{0.5}$, image tokens form equal contiguous camera blocks, and we use the first, top-camera block. For GR00T, we use the 88 tokens identified by the processor as the coordinate-consistent \texttt{cam\_high} block. Keys are measured after positional rotation and values after their value projection. The fourth GR00T VLSA block is not revised and serves only as the observer for the NTL before block 3.

\begin{figure*}[t]
    \centering
    \includegraphics[width=\textwidth]{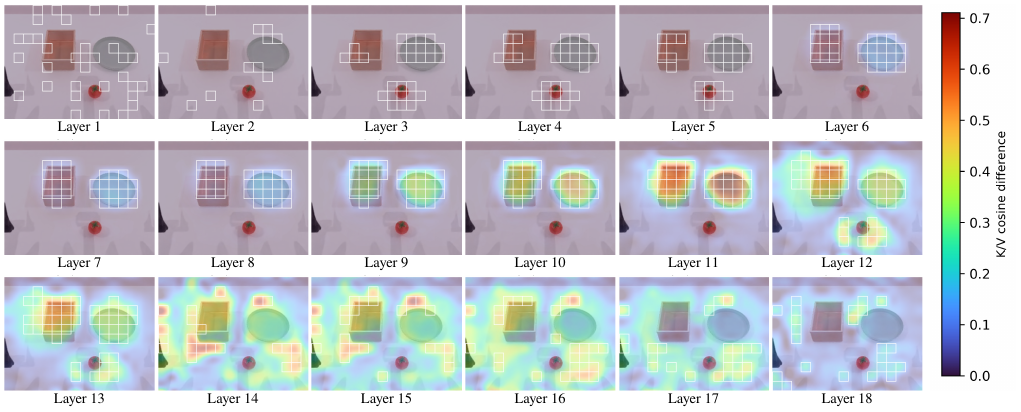}
    \caption{Layer-wise teacher-induced visual Key and Value differences and tokens retained by TGVA. Warmer colors indicate larger differences between satisfying and excluded instructions, and white boxes mark the adaptively selected tokens. Middle-layer selections concentrate on task-relevant objects, whereas deeper-layer selections become spatially diffuse.}
    \label{fig:tgva-selection}
\end{figure*}

For a candidate visual token $i$, we average cosine distance over KV heads and over keys and values:
\begin{equation}
D_i^l(A,B)=\frac{1}{2H_{\mathrm{kv}}}
\sum_{h=1}^{H_{\mathrm{kv}}}
\left[
1-\cos(\mathbf{k}_{A,i,h}^l,\mathbf{k}_{B,i,h}^l)
+1-\cos(\mathbf{v}_{A,i,h}^l,\mathbf{v}_{B,i,h}^l)
\right].
\label{eq:appendix-kv-distance}
\end{equation}
For sample $b$, we compute $d_{b,i}^l=D_i^l(\mathcal{O}_{T,b}^{a},\mathcal{O}_{T,b}^{e})$ for every top-camera token. Let $\mu_b^l$ and $s_b^l$ be the mean and standard deviation of these $N$ scores at layer $l$, and let $k=\lceil0.1N\rceil$. We select the token indices
\begin{equation}
\Phi_b^l=\{i\mid d_{b,i}^l>\mu_b^l+s_b^l\},\qquad
\Psi_b^l=
\begin{cases}
\Phi_b^l, & |\Phi_b^l|\geq k,\\
\operatorname{TopK}_k(d_{b,:}^l), & |\Phi_b^l|<k,
\end{cases}
\label{eq:appendix-token-selection}
\end{equation}
where $\operatorname{TopK}_k$ returns the indices of the $k$ largest scores. The fallback prevents weak teacher contrast from producing an empty or nearly empty training signal. Figure~\ref{fig:tgva-selection} visualizes the resulting teacher-induced Key and Value differences and retained visual tokens across layers. Within the selected set, weights are proportional to the teacher contrast,
\begin{equation}
w_{b,i}^l=
\frac{\mathbb{I}[i\in\Psi_b^l](d_{b,i}^l+10^{-6})}
{\sum_j\mathbb{I}[j\in\Psi_b^l](d_{b,j}^l+10^{-6})}.
\label{eq:appendix-token-weight}
\end{equation}
The set, distances, and normalized weights are detached; consequently, neither teacher receives gradients and the model cannot reduce the loss by manipulating which tokens are selected.

The excluded teacher is used only to expose instruction-sensitive locations. The alignment target itself is the satisfying teacher. At every selected layer, the equivalence loss aligns the negated student visual Key and Value with $\mathcal{O}_{T}^{a}$ on the selected tokens. In parallel, the semantic loss aligns the negated resolver output $\mathbf{r}^{n,l}$ with the detached affirmative student content $\mathbf{c}^{a,l}$, and the gate loss classifies the $B$ affirmative and $B$ negated student inputs with targets zero and one, respectively. These are precisely the three terms in Eq.~\ref{eq:objectives}. Losses are averaged uniformly over insertion sites. Gradients pass through the negated student Key and Value and the frozen VLM operations into the NTLs, but the original VLA weights, both teachers, and the teacher-derived selection remain fixed. Thus, TGVA supplies object-level spatial guidance without object masks, action targets, or trainable teacher networks.

\subsection{Optimization}
\paragraph{Objective and optimizer.}
Only the NTL parameters are passed to AdamW; all pretrained VLA parameters remain frozen throughout post-training. We optimize Eq.~\ref{eq:objectives} with $\beta_1=0.9$, $\beta_2=0.95$, $\epsilon=10^{-8}$, weight decay $10^{-10}$, and clip the global norm of all trainable NTL gradients to 1.0. These settings are shared by all three backbones.

\paragraph{Learning-rate schedule.}
Let $T$ be the total number of optimization steps and $W$ the number of warmup steps. The learning rate at step $t$ is
\begin{equation}
\eta_t=
\begin{cases}
\eta_0+(\eta_{\max}-\eta_0)t/W, & t<W,\\
\eta_{\min}+\frac{1}{2}(\eta_{\max}-\eta_{\min})
\left[1+\cos\!\left(\pi\frac{t-W}{T-W}\right)\right], & t\geq W,
\end{cases}
\end{equation}
where $\eta_{\max}=10^{-4}$, $\eta_{\min}=10^{-5}$, and $\eta_0=\eta_{\max}/(W+1)$. For simulation, we train for $T=20{,}000$ steps with $W=2{,}000$ warmup steps. Real-world $\pi_{0.5}$ adaptation uses the same optimizer and endpoint learning rates but trains for $T=5{,}000$ steps with $W=500$.

\paragraph{Distributed batching.}
The reported batch size of 4 is the \emph{global} number of independent scene--instruction examples per update. Under distributed data parallel training, it is divided evenly across workers, giving a per-worker batch size of $4/N$ for $N$ workers; no gradient accumulation is used. The affirmative and negated student branches and the two frozen teacher branches are generated internally from each example and therefore do not change this optimization-batch definition.

\paragraph{Precision.}
Model forward passes use bfloat16, while cosine distances, gate binary cross-entropy, loss aggregation, and other reduction-sensitive calculations are evaluated in FP32. Table~\ref{tab:appendix-training-config} summarizes the principal model-specific settings.

\begin{table*}[t]
\caption{NegaAlign post-training configurations. Layer indices are one-based.}
\label{tab:appendix-training-config}
\centering
\small
\setlength{\tabcolsep}{5pt}
\renewcommand{\arraystretch}{1.12}
\begin{tabular}{lcccccc}
\toprule
Backbone & Insertion sites & Trainable param. & Steps & Batch size & Warmup & Precision \\
\midrule
GR00T & VLSA 1--3 & 4.35M & 20,000 & 4 & 2,000 & bfloat16 \\
$\pi_0$ & VLM 6--13 & 11.6M & 20,000 & 4 & 2,000 & bfloat16 \\
$\pi_{0.5}$ & VLM 6--13 & 11.6M & 20,000 & 4 & 2,000 & bfloat16 \\
$\pi_{0.5}$ (real) & VLM 6--13 & 11.6M & 5,000 & 4 & 500 & bfloat16 \\
\bottomrule
\end{tabular}
\end{table*}

\section{Data Collection and Post-Training Pipeline}
\label{sec:data_pipeline}

\subsection{Simulation Data Collection and Composition}
\label{sec:simulation_data}
We construct the simulation data in RoboTwin 2.0 using SAPIEN and the dual-arm ALOHA-AgileX embodiment. The benchmark contains two scenes for each of five negation categories---object, attribute, comparison, goal, and complex---giving 10 scenarios in total. Across their affirmative and negated semantic variants, the environment defines 38 language-conditioned tasks. Objects, attributes, relative relations, and target placements are varied through seeded scene initialization.

Demonstrations are generated by the task-specific scripted expert using motion-planned grasping and placement primitives. A trajectory is retained only when motion planning succeeds and the binary success predicate of task is met. Each retained episode stores the seed, instruction, joint trajectory, end-effector state, and synchronized RGB observations. The raw HDF5 files contain four $320\times240$ camera streams (front, head, left wrist, and right wrist) and a 14-dimensional action vector ordered as the six left-arm joints, left gripper, six right-arm joints, and right gripper. The models use the head and two wrist views after conversion to the common LeRobot format.

The collected training corpus contains 2,360 episodes and 402,241 frames at 30 FPS. It comprises 38 task groups: 25 affirmative groups with 1,560 episodes and 13 explicitly negated groups with 800 episodes. Groups contain either 40 or 80 successful episodes. The explicit negated trajectories are used only by the fully supervised Pos.+Neg. baseline. Stage~1 affirmative training and NegaAlign Stage~2 do not use their observations, states, or actions.

\begin{figure*}[p]
    \centering
    \includegraphics[width=0.65\textwidth]{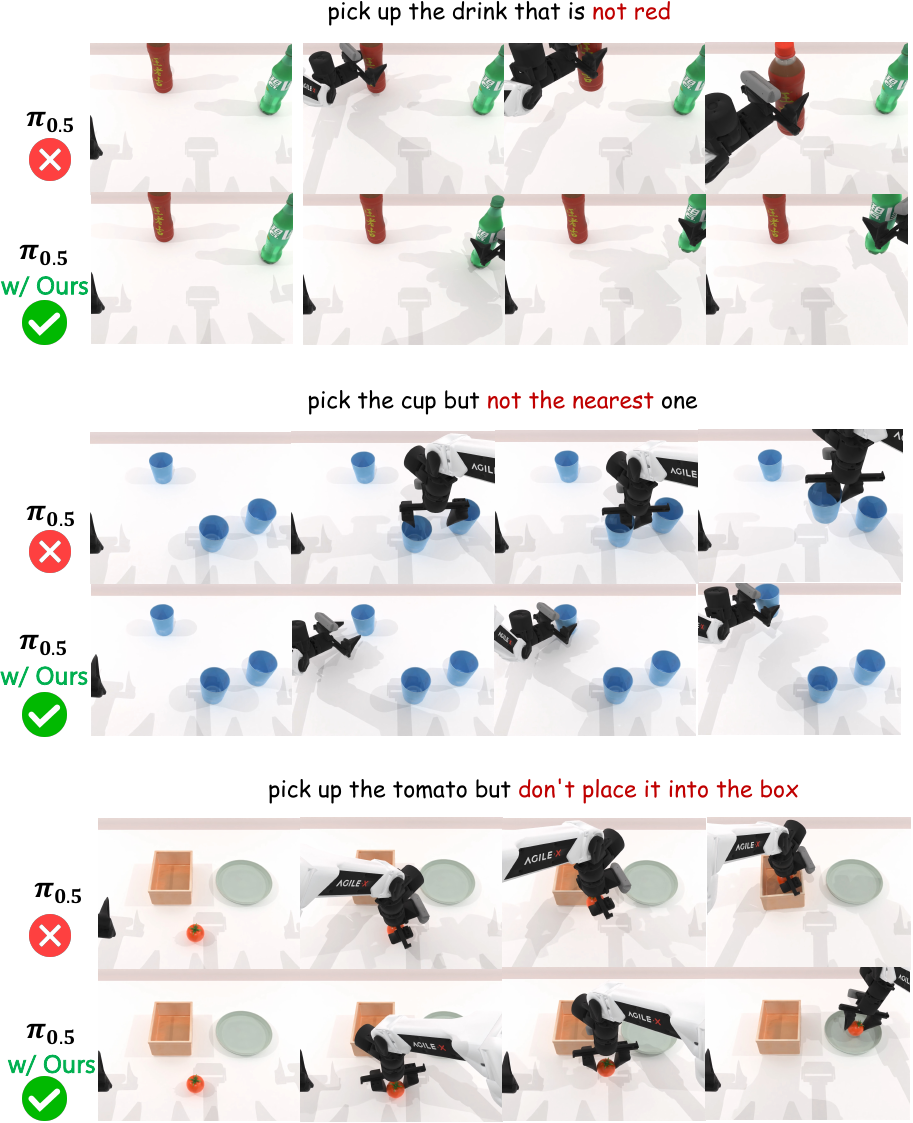}\\
    \includegraphics[width=0.65\textwidth]{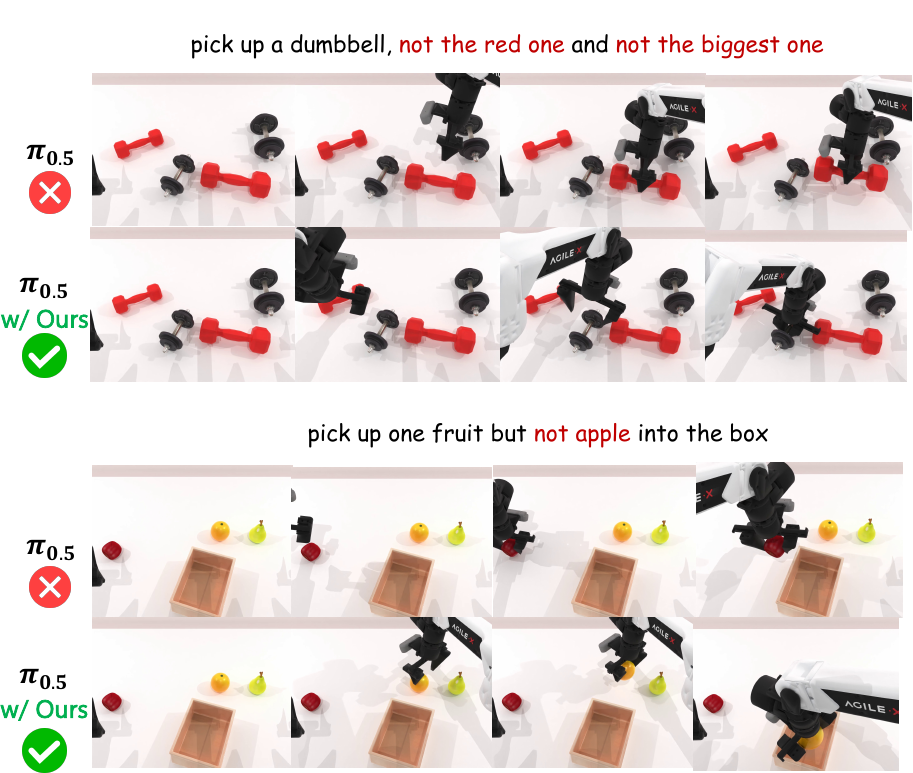}
    \caption{Qualitative evaluation results of affirmative-only $\pi_{0.5}$ and NegaAlign-adapted $\pi_{0.5}$ on five representative NegaBench scenarios. For each negated instruction, the top row shows the affirmative-only baseline violating the constraint, while the bottom row shows NegaAlign satisfying it. The columns show successive stages of each rollout.}
    \label{fig:appendix-simulation-evaluation}
\end{figure*}

\subsection{Real-World Data Composition}
\label{sec:realworld_data}
The real-world dataset is collected on a UFACTORY xArm6 and contains 500 affirmative demonstrations, totaling 208,070 frames (approximately 1.93 hours) at 30 FPS. It covers 10 tasks with 50 episodes per task. The tasks form five semantic contrast pairs: spoon versus knife, separated versus mixed food, shirt versus socks, leaving versus storing a diamond, and ceramic versus stainless-steel bowls. No negated robot demonstration is included in this dataset.

The data follow the LeRobot v2.1 organization, with one Parquet trajectory and two synchronized MP4 streams per episode. The overhead camera is recorded at $1920\times1080$ and the wrist-mounted tool camera at $640\times480$. Each frame contains a 7-dimensional state and action vector ordered as 3-D position, 3-D rotation, and gripper position. The 10 tasks are balanced by episode count, although their durations differ: the dataset has a mean of 416.14 frames per episode, with lengths ranging from 229 to 1,027 frames. For Stage~1, all 500 affirmative episodes are available to train the base policy.

\begin{figure*}[p]
    \centering
    \includegraphics[width=0.62\textwidth]{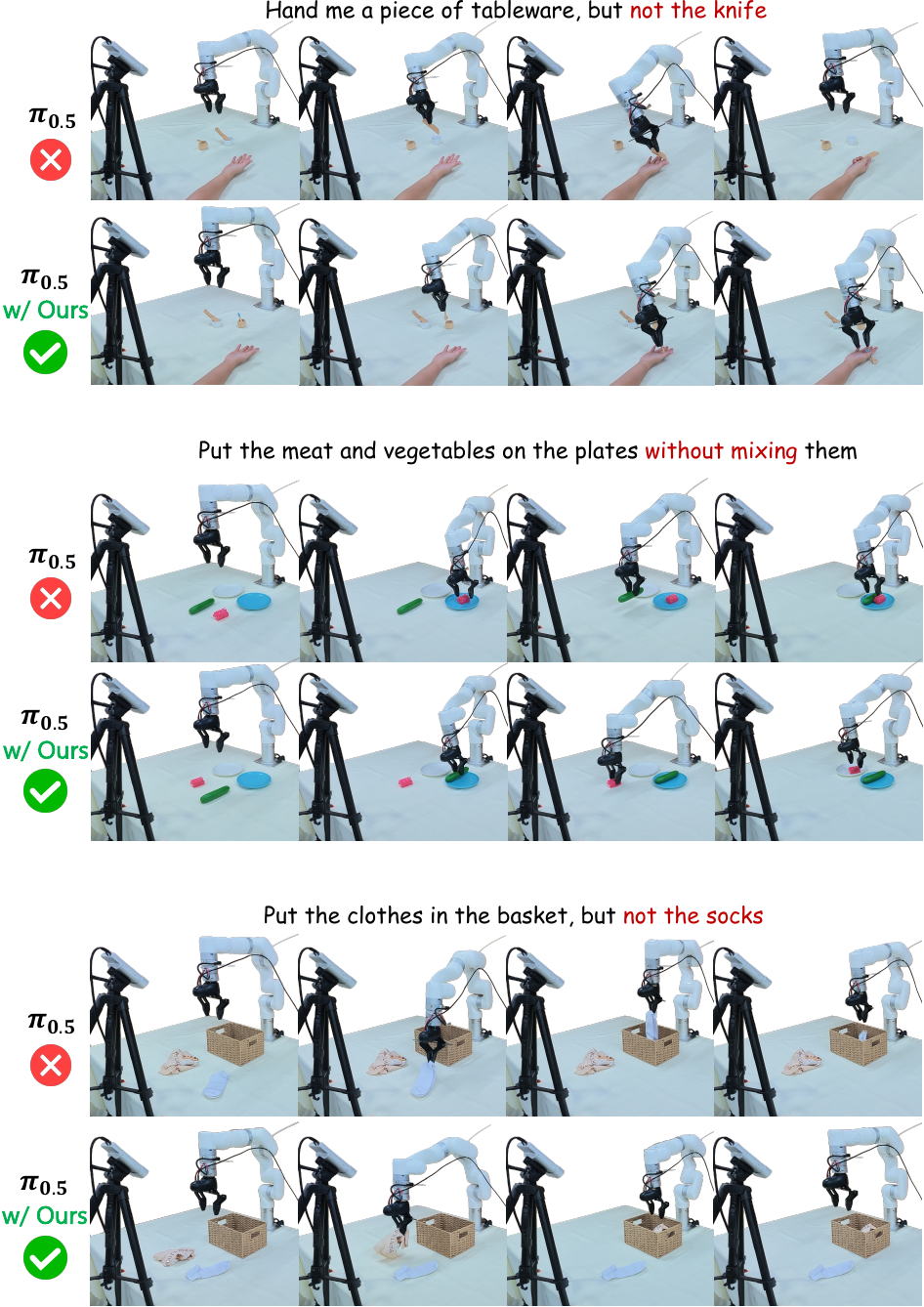}\\
    \includegraphics[width=0.62\textwidth]{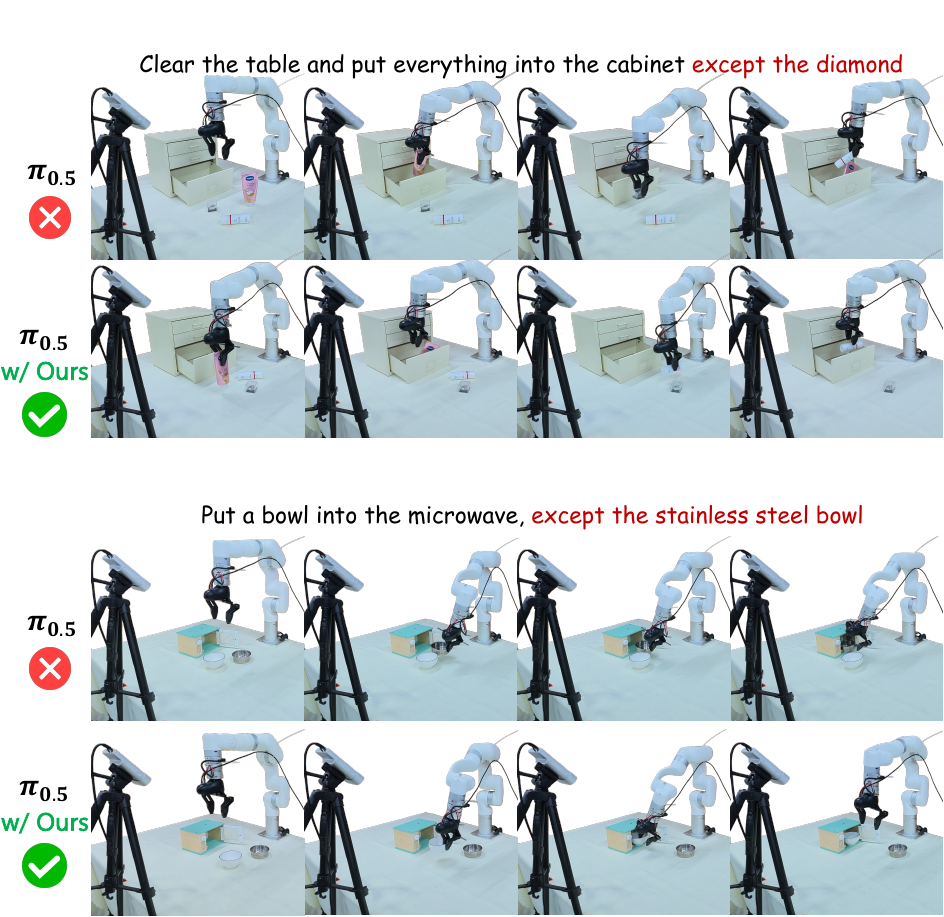}
    \caption{Qualitative evaluation results of affirmative-only $\pi_{0.5}$ and NegaAlign-adapted $\pi_{0.5}$ on five real-world xArm6 scenarios. For each negated instruction, the top row shows the affirmative-only baseline violating the constraint, while the bottom row shows NegaAlign satisfying it. The columns show successive stages of each rollout.}
    \label{fig:appendix-real-evaluation}
\end{figure*}

\subsection{Construction of Instruction Relations}
\label{sec:supervision_details}
We manually organize the available affirmative skills into the relation $\mathcal{R}_q=(\ell_q^n,\mathcal{A}_q,\ell_q^e)$ . The simulation configuration contains 13 relations, listed in Table~\ref{tab:simulation-instruction-relations}. Most have one satisfying interpretation; the two food-exclusion relations retain both valid alternatives so that the supervision does not impose an arbitrary single behavior.

\begin{table*}[t]
\caption{Complete instruction relations used for simulation post-training. For relations with multiple valid satisfying instructions, all alternatives are retained in $\mathcal{A}$.}
\label{tab:simulation-instruction-relations}
\centering
\footnotesize
\setlength{\tabcolsep}{5pt}
\renewcommand{\arraystretch}{1.08}
\begin{tabular}{@{}>{\raggedright\arraybackslash}p{0.12\textwidth}>{\raggedright\arraybackslash}p{0.82\textwidth}@{}}
\toprule
Type & Instruction relation \\
\midrule
Attribute & \textbf{Negated:} Pick up the drink that is not red.\newline
\textbf{Satisfying:} Pick up the drink that is green.\newline
\textbf{Excluded:} Pick up the drink that is red. \\
Attribute & \textbf{Negated:} Pick the cup that is not full.\newline
\textbf{Satisfying:} Pick the cup that is empty.\newline
\textbf{Excluded:} Pick the cup that is full. \\
\addlinespace[2pt]
Comparison & \textbf{Negated:} Pick the cup but not the nearest one.\newline
\textbf{Satisfying:} Pick the farthest cup.\newline
\textbf{Excluded:} Pick the nearest cup. \\
Comparison & \textbf{Negated:} Put one shoe in the shoebox, but do not choose the biggest one.\newline
\textbf{Satisfying:} Put one shoe in the shoebox; choose the smallest one.\newline
\textbf{Excluded:} Put one shoe in the shoebox; choose the biggest one. \\
\addlinespace[2pt]
Complex & \textbf{Negated:} Pick up the bottle but not the water, and place it on the plate but not the left one.\newline
\textbf{Satisfying:} Pick up the soy sauce and place it on the right plate.\newline
\textbf{Excluded:} Pick up the water and place it on the left plate. \\
Complex & \textbf{Negated:} Pick up the bottle but not the soy sauce, and place it on the plate but not the right one.\newline
\textbf{Satisfying:} Pick up the water and place it on the left plate.\newline
\textbf{Excluded:} Pick up the soy sauce and place it on the right plate. \\
Complex & \textbf{Negated:} Pick up a dumbbell, not the red one and not the biggest one.\newline
\textbf{Satisfying:} Pick up the black and smallest dumbbell.\newline
\textbf{Excluded:} Pick up the red and biggest dumbbell. \\
Complex & \textbf{Negated:} Pick up a dumbbell, not the black one and not the smallest one.\newline
\textbf{Satisfying:} Pick up the red and biggest dumbbell.\newline
\textbf{Excluded:} Pick up the black and smallest dumbbell. \\
\addlinespace[2pt]
Goal & \textbf{Negated:} Pick up the tomato but do not place it into the box.\newline
\textbf{Satisfying:} Pick up the tomato and place it on the plate.\newline
\textbf{Excluded:} Pick up the tomato and place it into the box. \\
Goal & \textbf{Negated:} Pick up the bread but do not place it on the table.\newline
\textbf{Satisfying:} Pick up the bread and place it on the plate.\newline
\textbf{Excluded:} Pick up the bread and place it on the table. \\
\addlinespace[2pt]
Object & \textbf{Negated:} Pick up one fruit but not the apple and put it into the box.\newline
\textbf{Satisfying:} Pick up the orange and put it into the box.\newline
\textbf{Excluded:} Pick up the apple and put it into the box. \\
Object & \textbf{Negated:} Place any food except the chips on the plate.\newline
\textbf{Satisfying:} Place the fries on the plate; or place the hamburger on the plate.\newline
\textbf{Excluded:} Place the chips on the plate. \\
Object & \textbf{Negated:} Place any food except the fries on the plate.\newline
\textbf{Satisfying:} Place the chips on the plate; or place the hamburger on the plate.\newline
\textbf{Excluded:} Place the fries on the plate. \\
\bottomrule
\end{tabular}
\end{table*}

For real-world adaptation, the 10 affirmative tasks yield the five relations in Table~\ref{tab:real-instruction-relations}. Each negated instruction is authored to preserve the common task while excluding one member of its affirmative pair. Consequently, the relation changes skill selection but does not require a new low-level manipulation primitive.

\begin{table*}[t]
\caption{Complete instruction relations used for real-world post-training.}
\label{tab:real-instruction-relations}
\centering
\footnotesize
\setlength{\tabcolsep}{5pt}
\renewcommand{\arraystretch}{1.1}
\begin{tabular}{@{}>{\raggedright\arraybackslash}p{0.12\textwidth}>{\raggedright\arraybackslash}p{0.82\textwidth}@{}}
\toprule
Scenario & Instruction relation \\
\midrule
Dining & \textbf{Negated:} Hand me a piece of tableware, but not the knife.\newline
\textbf{Satisfying:} Hand me the spoon.\newline
\textbf{Excluded:} Hand me the knife. \\
Sorting & \textbf{Negated:} Put the meat and vegetables on the plates without mixing them.\newline
\textbf{Satisfying:} Put the meat and vegetable on separate plates.\newline
\textbf{Excluded:} Put the meat and vegetable on the plates, mixing them together. \\
Laundry & \textbf{Negated:} Put the clothes in the basket, but not the socks.\newline
\textbf{Satisfying:} Put the shirt in the basket.\newline
\textbf{Excluded:} Put the socks in the basket. \\
Storage & \textbf{Negated:} Clear the table and put everything into the cabinet except the diamond.\newline
\textbf{Satisfying:} Clear the table and put everything into the cabinet, leaving the diamond on the table.\newline
\textbf{Excluded:} Clear the table and put everything into the cabinet, including the diamond. \\
Cooking & \textbf{Negated:} Put a bowl into the microwave, except the stainless-steel bowl.\newline
\textbf{Satisfying:} Put the ceramic bowl into the microwave.\newline
\textbf{Excluded:} Put the stainless-steel bowl into the microwave. \\
\bottomrule
\end{tabular}
\end{table*}

Before training, whitespace is normalized and all members of a relation are required to be nonempty and distinct. The episode manifest verifies that every satisfying and excluded affirmative instruction exists in the affirmative dataset and, critically, that no episode carries the negated instruction. The simulation manifest then loads all 840 episodes associated with 14 unique satisfying prompts; the real-world manifest loads 250 episodes associated with five satisfying prompts. Excluded-task episodes are checked for availability but are not loaded into the Stage~2 sampling pool. They define a language contrast, not an additional source of visual observations.

\subsection{End-to-End Post-Training Pipeline}
\label{sec:posttraining_pipeline}
The Stage~2 loader first samples a relation uniformly, then samples a satisfying instruction uniformly from its set, and finally samples a frame uniformly from all episodes associated with that instruction. This hierarchical procedure prevents relations with more satisfying alternatives, longer trajectories, or more frames from dominating the objective. Simulation observations are repacked as the three ALOHA policy views and a 14-dimensional state; real-world observations are repacked as overhead/tool views and a 7-dimensional state. We apply the same input transforms and quantile normalization used by the corresponding Stage~1 policy, resize images to $224\times224$, tokenize the prompt, and explicitly mark only the instruction-token span. Actions may be read by the common dataset interface, but they are never supplied as a Stage~2 target and do not contribute to any NegaAlign loss.

For every sampled $(\mathbf{o},\mathbf{s},\ell^a)$, the pipeline retrieves its relation and re-tokenizes the same observation--state pair with $\ell^n$ and $\ell^e$. It constructs a joint $2B$ student batch containing $(\ell^a,\ell^n)$ and two $B$-sized, no-gradient teacher branches containing $(\ell^a,\ell^e)$. The affirmative and negated student samples receive polarity targets 0 and 1, respectively. The satisfying teacher provides the semantic and visual destination, whereas the difference between satisfying and excluded teachers selects the TGVA visual tokens. The three losses in Eq.~\ref{eq:objectives} update only the NTL parameters. The teacher branches and all original VLA components remain frozen, and the action expert is not executed during post-training. Architecture and optimizer details are provided in Appendix~\ref{sec:training_details}.

At inference time, relation lookup and both teacher branches are removed. A single observation, state, and user instruction pass through the frozen VLA augmented with the trained NTLs, after which the original action generator produces the action chunk. Thus, the additional training structure has no teacher-model cost at deployment.

\section{Evaluation Protocol and Detailed Results}
\label{sec:evaluation_details}

\subsection{NegaBench Evaluation Protocol}
\label{sec:simulation_eval_protocol}
All simulation policies are evaluated through the same RoboTwin client--server pipeline. The client owns environment initialization, instruction selection, rollout, success checking, and result logging, while a model-specific server only converts observations to an action chunk. This separation permits each VLA to use its own software environment without changing the benchmark logic.

For every evaluated task variant, we run 100 episodes. Candidate initialization seeds are first executed by the scripted expert of task; a seed is admitted only when motion planning succeeds and the expert satisfies the same task-specific success predicate used for policy evaluation. The resulting task/config-specific seed list is cached and reused across policies. We fix the task manifest, task configuration, seed group, checkpoint, normalization statistics, and action execution horizon for each comparison.

At each closed-loop step, the policy receives the instruction, a head-view image, two wrist-view images, and its model-specific proprioceptive state. The $\pi_0$, $\pi_{0.5}$, and GR00T adapters execute 14-dimensional dual-arm joint-position actions. APT instead executes 16-dimensional dual-arm end-effector pose and gripper actions. A predicted chunk is truncated to the minimum of the configured execution horizon (32), the returned chunk length, and the remaining episode budget. The environment checks success throughout low-level action execution and terminates on the first successful state or after 300 high-level actions.

Each episode produces a binary outcome. For affirmative instructions, success requires completing the requested manipulation. For negated instructions, it additionally requires respecting every exclusion constraint. For task $k$ with $N=100$ episodes, we report
\begin{equation}
    \mathrm{SR}_k=\frac{1}{N}\sum_{i=1}^{N}\mathbb{I}[\text{episode }i\text{ succeeds}].
\end{equation}
Category and overall values in the main paper are macro-averages of the corresponding task-level success rates, so tasks with longer trajectories do not receive additional weight.
Representative evaluation rollouts of affirmative-only $\pi_{0.5}$ and NegaAlign-adapted $\pi_{0.5}$ across the five NegaBench categories are shown in Figure~\ref{fig:appendix-simulation-evaluation}.

\subsection{Real-World Evaluation Protocol}
\label{sec:realworld_eval_details}
We evaluate $\pi_{0.5}$ and its NegaAlign-adapted counterpart on the five xArm6 scenarios in Table~\ref{tab:real-instruction-relations}. The negated constraints respectively exclude a knife, mixed food placement, socks, moving a diamond, and a stainless-steel bowl. A trial is successful only when the shared manipulation goal is completed and the excluded object or outcome is avoided.

Each negated instruction is evaluated for 50 trials. Each scenario also has two affirmative counterparts; we run 25 trials for each and pool their outcomes, producing the same 50-trial denominator for affirmative and negated evaluation. We report both the integer success count and success rate for every scenario, followed by the unweighted mean across the five scenarios. This pairing separates failure to perform the underlying manipulation from failure to obey the negated constraint.
Representative evaluation rollouts of affirmative-only $\pi_{0.5}$ and NegaAlign-adapted $\pi_{0.5}$ across the five real-world scenarios are shown in Figure~\ref{fig:appendix-real-evaluation}.

\subsection{Detailed Simulation Results}
\label{sec:detailed_simulation_results}
Table~\ref{tab:detailed-simulation-results} reports the task-level success rates underlying Table~\ref{tab:Comparison}. When a task contains multiple instruction variants, we first average those variants within the task; category and overall scores are then macro-averaged across tasks. The final row of each model block therefore follows exactly the aggregation protocol used in the main paper.

\begin{table*}[t]
\caption{Detailed NegaBench success rates (\%). Pos. FT uses only affirmative trajectories, Pos.+Neg. FT additionally uses explicit negated trajectories, and NegaAlign uses affirmative trajectories only during post-training.}
\label{tab:detailed-simulation-results}
\centering
\scriptsize
\setlength{\tabcolsep}{4pt}
\renewcommand{\arraystretch}{1.03}
\begin{tabular}{lllrr>{\columncolor{red!7}}r>{\columncolor{red!7}}r>{\columncolor{green!9}}r>{\columncolor{green!9}}r}
\toprule
\multirow{2}{*}{Model} & \multirow{2}{*}{Category} & \multirow{2}{*}{Task}
& \multicolumn{2}{c}{Pos. FT}
& \multicolumn{2}{>{\columncolor{red!7}}c}{Pos.+Neg. FT}
& \multicolumn{2}{>{\columncolor{green!9}}c}{w/ Ours} \\
\cmidrule(lr){4-5}\cmidrule(lr){6-7}\cmidrule(lr){8-9}
& & & Pos. & Neg. & Pos. & Neg. & Pos. & Neg. \\
\midrule
\multirow{11}{*}{$\pi_{0.5}$}
& \multirow{2}{*}{Object} & 1 & 88.50 & 2.00 & 83.00 & 71.00 & 83.00 & 79.00 \\
& & 2 & 94.67 & 5.00 & 88.33 & 84.50 & 92.33 & 76.00 \\
& \multirow{2}{*}{Attribute} & 1 & 99.00 & 1.00 & 97.50 & 98.00 & 99.00 & 97.00 \\
& & 2 & 77.00 & 0.00 & 83.50 & 88.00 & 84.50 & 92.00 \\
& \multirow{2}{*}{Comparison} & 1 & 96.00 & 12.00 & 84.00 & 84.00 & 94.00 & 98.00 \\
& & 2 & 82.00 & 2.00 & 87.50 & 78.00 & 85.00 & 58.00 \\
& \multirow{2}{*}{Goal} & 1 & 92.00 & 0.00 & 93.50 & 85.00 & 89.50 & 87.00 \\
& & 2 & 100.00 & 0.00 & 100.00 & 100.00 & 99.00 & 100.00 \\
& \multirow{2}{*}{Complex} & 1 & 99.25 & 4.00 & 98.25 & 99.00 & 100.00 & 98.50 \\
& & 2 & 99.25 & 0.00 & 98.50 & 98.50 & 98.75 & 99.00 \\
& \multicolumn{2}{l}{Average} & 92.77 & 2.60 & 91.41 & 88.60 & 92.51 & 88.45 \\
\midrule
\multirow{11}{*}{GR00T}
& \multirow{2}{*}{Object} & 1 & 51.50 & 2.00 & 55.50 & 36.00 & 52.00 & 26.00 \\
& & 2 & 68.67 & 0.50 & 66.33 & 52.00 & 67.67 & 23.50 \\
& \multirow{2}{*}{Attribute} & 1 & 84.50 & 0.00 & 83.50 & 90.00 & 87.50 & 77.00 \\
& & 2 & 43.00 & 0.00 & 36.50 & 41.00 & 50.00 & 59.00 \\
& \multirow{2}{*}{Comparison} & 1 & 82.50 & 8.00 & 92.00 & 82.00 & 88.00 & 84.00 \\
& & 2 & 57.50 & 0.00 & 51.00 & 48.00 & 57.50 & 39.00 \\
& \multirow{2}{*}{Goal} & 1 & 53.50 & 0.00 & 85.50 & 59.00 & 56.00 & 79.00 \\
& & 2 & 96.50 & 0.00 & 97.00 & 96.00 & 96.00 & 96.00 \\
& \multirow{2}{*}{Complex} & 1 & 79.50 & 0.00 & 90.50 & 90.00 & 79.75 & 33.00 \\
& & 2 & 93.75 & 0.00 & 96.75 & 94.50 & 94.50 & 61.00 \\
& \multicolumn{2}{l}{Average} & 71.09 & 1.05 & 75.46 & 68.85 & 72.89 & 57.75 \\
\midrule
\multirow{11}{*}{$\pi_0$}
& \multirow{2}{*}{Object} & 1 & 40.50 & 14.00 & 30.50 & 27.00 & 39.50 & 35.00 \\
& & 2 & 21.00 & 15.50 & 29.33 & 26.50 & 19.33 & 14.50 \\
& \multirow{2}{*}{Attribute} & 1 & 65.00 & 36.00 & 92.00 & 96.00 & 64.00 & 58.00 \\
& & 2 & 52.00 & 52.00 & 57.50 & 70.00 & 55.00 & 61.00 \\
& \multirow{2}{*}{Comparison} & 1 & 90.00 & 15.00 & 87.00 & 87.00 & 92.00 & 88.00 \\
& & 2 & 61.00 & 21.00 & 68.50 & 58.00 & 61.00 & 31.00 \\
& \multirow{2}{*}{Goal} & 1 & 92.50 & 0.00 & 98.00 & 88.00 & 92.00 & 93.00 \\
& & 2 & 100.00 & 0.00 & 99.00 & 100.00 & 100.00 & 100.00 \\
& \multirow{2}{*}{Complex} & 1 & 99.50 & 0.00 & 96.25 & 97.50 & 99.25 & 95.50 \\
& & 2 & 60.50 & 0.50 & 58.75 & 62.50 & 62.00 & 57.00 \\
& \multicolumn{2}{l}{Average} & 68.20 & 15.40 & 71.68 & 71.25 & 68.41 & 63.30 \\
\bottomrule
\end{tabular}
\end{table*}
\FloatBarrier

Affirmative-only fine-tuning fails systematically on negated instructions: the task-level negated success rates are at most 12.00\% for $\pi_{0.5}$ and 8.00\% for GR00T, while $\pi_0$ remains at or below 52.00\% on every task. NegaAlign improves negated success on all 30 backbone--task combinations without using negated trajectories. The gains are not confined to a single linguistic pattern: $\pi_{0.5}$ reaches at least 76.00\% on every object task, at least 92.00\% on every attribute task, and at least 98.50\% on every complex task. Its overall negated score of 88.45\% nearly matches the 88.60\% obtained with explicit negated demonstrations.

The task breakdown also identifies the remaining backbone-dependent failure modes. GR00T is weakest on object exclusion and on the first complex task, whereas $\pi_0$ retains larger gaps on object and complex constraints. Despite these differences, affirmative performance is preserved: relative to Pos. FT, NegaAlign changes the overall affirmative score by $+1.80$, $+0.21$, and $-0.26$ percentage points for GR00T, $\pi_0$, and $\pi_{0.5}$, respectively. The improvement therefore reflects constraint-sensitive behavior rather than a trade-off that sacrifices the original manipulation skills.

\subsection{Detailed Real-World Results}
\label{sec:detailed_realworld_results}
Table~\ref{tab:detailed-real-results} reports the success count for every real-world instruction. Each negated instruction uses 50 trials, while each of its satisfying and excluded affirmative controls uses 25 trials. The pooled rows exactly recover the affirmative and negated success rates reported in Table~\ref{tab:real-world}.

\begin{table*}[!t]
\caption{Instruction-level real-world success counts. The satisfying and excluded affirmative controls are evaluated separately and pooled only in the final Pos. row.}
\label{tab:detailed-real-results}
\centering
\footnotesize
\setlength{\tabcolsep}{5pt}
\renewcommand{\arraystretch}{1.06}
\begin{tabular}{@{}>{\raggedright\arraybackslash}p{0.10\textwidth}>{\raggedright\arraybackslash}p{0.575\textwidth}>{\columncolor{baseCell}\centering\arraybackslash}p{0.105\textwidth}>{\columncolor{oursCell}\centering\arraybackslash}p{0.145\textwidth}@{}}
\toprule
Scenario & Instruction
& \multicolumn{1}{c}{$\pi_{0.5}$}
& \multicolumn{1}{c}{\shortstack{$\pi_{0.5}$ w/ Ours}} \\
\midrule
Sorting
& \textbf{Negated:}\newline Put the meat and vegetables on the plates without mixing them. & 9/50 & 40/50 \\
& \textbf{Satisfying:}\newline Put the meat and vegetable on separate plates. & 22/25 & 23/25 \\
& \textbf{Excluded:}\newline Put the meat and vegetable on the plates, mixing them together. & 24/25 & 19/25 \\
\addlinespace[2pt]
Laundry
& \textbf{Negated:}\newline Put the clothes in the basket, but not the socks. & 4/50 & 47/50 \\
& \textbf{Satisfying:}\newline Put the shirt in the basket. & 23/25 & 25/25 \\
& \textbf{Excluded:}\newline Put the socks in the basket. & 23/25 & 24/25 \\
\addlinespace[2pt]
Storage
& \textbf{Negated:}\newline Clear the table and put everything into the cabinet except the diamond. & 1/50 & 50/50 \\
& \textbf{Satisfying:}\newline Clear the table and put everything into the cabinet, leaving the diamond on the table. & 22/25 & 22/25 \\
& \textbf{Excluded:}\newline Clear the table and put everything into the cabinet, including the diamond. & 23/25 & 22/25 \\
\addlinespace[2pt]
Cooking
& \textbf{Negated:}\newline Put a bowl into the microwave, except the stainless-steel bowl. & 0/50 & 48/50 \\
& \textbf{Satisfying:}\newline Put the ceramic bowl into the microwave. & 24/25 & 25/25 \\
& \textbf{Excluded:}\newline Put the stainless-steel bowl into the microwave. & 25/25 & 23/25 \\
\addlinespace[2pt]
Dining
& \textbf{Negated:}\newline Hand me a piece of tableware, but not the knife. & 17/50 & 37/50 \\
& \textbf{Satisfying:}\newline Hand me the spoon. & 24/25 & 23/25 \\
& \textbf{Excluded:}\newline Hand me the knife. & 19/25 & 21/25 \\
\midrule
All
& \textbf{Negated:}\newline All five negated instructions & 31/250 (12.4\%) & 222/250 (88.8\%) \\
& \textbf{Affirmative:}\newline All ten affirmative instructions & 229/250 (91.6\%) & 227/250 (90.8\%) \\
\bottomrule
\end{tabular}
\end{table*}
\FloatBarrier

The affirmative-only policy completes 229 of 250 affirmative trials but only 31 of 250 negated trials, confirming that task competence alone does not yield exclusion-aware behavior. NegaAlign raises negated success in every scenario: Sorting improves from 9/50 to 40/50, Laundry from 4/50 to 47/50, Storage from 1/50 to 50/50, Cooking from 0/50 to 48/50, and Dining from 17/50 to 37/50. The gain therefore spans constraints on objects, attributes, and outcomes rather than being driven by one task family.

At the same time, the adapted policy succeeds in 227 of 250 affirmative trials, only two fewer than the base policy. In particular, it remains able to execute the explicitly excluded affirmative alternative when that behavior is directly requested, such as placing the stainless-steel bowl in the microwave or handing over the knife. This rules out a trivial strategy that globally suppresses the excluded object and shows that the behavioral change is conditioned on the negated instruction. Dining remains the most challenging negated scenario at 37/50, suggesting that open-ended object selection leaves more residual ambiguity than the more structured placement tasks.
\end{document}